\documentclass{article}
\usepackage{iclr2027_conference,times}

\newcommand{\indicator}{\mathbf{1}}

\usepackage{amsmath,amssymb,booktabs,graphicx,microtype,multirow,tabularx,xcolor,hyperref,url}
\hypersetup{hidelinks}
\graphicspath{{figures/}}
\newcommand{\finalonly}{\textsc{final-only}}
\newcommand{\allprefix}{\textsc{all-prefix}}
\newcommand{\sampledprefix}{\textsc{sampled-prefix}}
\newcommand{\matchedexit}{\textsc{matched random exit}}
\newcommand{\progressive}{\textsc{progressive}}

\newcommand{\double}{\textsc{double}}

\title{Learning to Compute Is Not Learning to Retain\\or Reuse in Looped Transformers}
\author{%
\textbf{Peilong Zhou$^{1,2,3,*}$ \quad Hongtao Zhang$^{2,*}$ \quad Yulong Huang$^{4}$}\\
\textbf{Cangyuan Li$^{1,2}$ \quad Mingyu Chen$^{2}$ \quad Ying Wang$^{1,2,\dagger}$}\\[0.5em]
\normalfont\small $^{1}$SKLP, Institute of Computing Technology, Chinese Academy of Sciences\\
\normalfont\small $^{2}$University of Chinese Academy of Sciences\\
\normalfont\small $^{3}$School of Advanced Interdisciplinary Sciences\\
\normalfont\small $^{4}$The Hong Kong University of Science and Technology (Guangzhou)\\[0.4em]
\normalfont\small $^{*}$Equal contribution.\quad $^{\dagger}$Corresponding author.\\
\normalfont\small \texttt{wangying2009@ict.ac.cn}
}

\iclrfinalcopy
\hypersetup{
  pdftitle={Learning to Compute Is Not Learning to Retain or Reuse in Looped Transformers},
  pdfauthor={Peilong Zhou, Hongtao Zhang, Yulong Huang, Cangyuan Li, Mingyu Chen, Ying Wang},
  pdfsubject={Preprint}
}
\begin{document}
\maketitle
\lhead{Preprint}
\begin{abstract}
Looped transformers compute by applying one block repeatedly, so they can run
for more loops than they were trained with. Recurrent reasoners are known to
overthink: iterating past the trained depth can turn a correct answer into a
wrong one. Such a late error can reflect failed computation, loss of
retention, or a state that later computation cannot reuse. We separate these cases in in-context
Mealy-machine execution, where a simulator supplies intermediate targets and
exact counterfactuals for hidden-state interventions. Models trained to answer
at loop~32 are fully correct there but keep only 22\% of these answers at
loop~64. Retention depends on what training asks for: supervising one randomly
chosen position per example, instead of always the last, keeps 99.8\%, and
randomizing the number of training loops also helps; truncating gradients, as
in progressive training against overthinking, does not. The computation itself is
exact: position~$k$ is computed at loop~$k$, and swapping its loop-$k$ hidden
vector for the one from another input redirects every later position to that
input's exact continuation (237/237 pairs in each of ten models). Retention,
however, is not reuse. Even in models that keep nearly all answers, holding
each position~$k$ fixed from loop~$k$ drops accuracy to 4--6\%, whereas
holding it from loop~$k+4$ is harmless. A released recurrent maze network
trained to avoid overthinking behaves the same way. Learning to compute is thus
not learning to retain, and avoiding overthinking does not make a state
reusable when its answer first appears.
\end{abstract}

\section{Introduction}

Looped transformers turn recurrent depth into a computation budget: one block
is applied repeatedly, so a model can run for more loops than it was trained
with \citep{dehghani2019universal,giannou2023looped,geiping2025scaling}.
Whether those extra loops help depends on what happens to an answer after it
has been computed.

Empirically, running past the trained depth can hurt. Recurrent networks
trained on small instances can solve larger ones by iterating longer
\citep{schwarzschild2021learn}, but further iteration can also turn a correct
answer into a wrong one. \citet{bansal2022endtoend} call this failure
\emph{overthinking} and prevent it with input recall and progressive
training, which applies the loss at a random depth and blocks gradients to
the loops before a random starting point. Later work relates extra
computation to path independence and shows that training and stopping
objectives change the learned trajectory
\citep{anil2022path,zhang2026recurrence,kuo2026stabilizing,popescu2026adaptive};
analysis of recurrent language models separates whether content is readable
from whether it is transported and can be steered \citep{wang2026jacobian}.
Much of this evidence tracks accuracy or readouts across loops. An
overthinking curve, however, can mean three different things:
the answer was never \emph{computed}, it was computed but not
\emph{retained}, or the state that carries it cannot be \emph{reused} by
later computation. Telling these apart requires knowing what a correct
alternative execution would be. We therefore ask: \emph{when a looped model
learns to compute an answer, what does it learn about retaining that answer
and reusing its internal state?}

In-context execution of Mealy machines makes these distinctions measurable
(Figure~\ref{fig:setting}). Each example gives a transition table and an
input sequence of length $H$. A simulator supplies the correct state and
output at every position, including after any counterfactual change, so an
intervention can be scored against a unique correct continuation rather
than by whether the output changed. We apply one shared transformer block for
$R$ loops and train only on the readout after the last loop, the \emph{exit};
the main models use $R=H$, one loop per input symbol, and compare objectives that differ only in which answers must be correct at
that exit: the last position (\finalonly), every position (\allprefix), or
one uniformly sampled position per example (\sampledprefix). Endpoint
accuracy requires both the state and the output at the last position to be
correct. In the models we analyze, position $k$ first becomes correct at or near
loop $k$; we call the diagonal $r=k$ the \emph{frontier}. Because the loop at
which each answer first appears is known, a state can be tested at exactly
that moment. We intervene on the recurrent
trajectory in three ways: replacing a hidden vector in a \emph{base run} with
the corresponding vector of a \emph{source run} on a different input prefix
\citep{geiger2024finding}, resetting a position to its own earlier vector,
and freezing each position at a chosen loop. At the frontier, one
position's hidden vector suffices to drive the rest of the execution, so
the same unit lets us compare what a state makes readable with what it lets
later computation do.

The three properties come apart. In the order of the paper, we find:
\begin{itemize}
\item \textbf{Retention is set by what training requires
(Section~\ref{sec:retention}).} On shared length-32 test inputs,
\finalonly\ models are 100\% correct at the trained exit but 22.3\% at
loop 64 (Figure~\ref{fig:behavior}). \allprefix\ gives 94.3\%, and
\sampledprefix\ gives 99.8\% with the same single supervised position per
example as \finalonly. At a matched expected budget, drawing the exit as
$R=H+U$ with $U$ uniform on $\{0,\ldots,8\}$, instead of fixing $R=H+4$,
raises loop-64 accuracy from 30.0\% to 85.6\%. Truncated gradients do not
help: our adaptation of progressive training learns length 32 in one of
three seeds, whereas a \matchedexit\ control with identical exits and full
gradients learns it in all three and keeps 97.9\% at loop 64.
\item \textbf{Computation is exact, but its answers are transient
(Section~\ref{sec:transport}).} Replacing one hidden vector at the frontier
redirects every later frontier readout to the exact counterfactual execution
in all 237 pairs whose counterfactual differs, in each of ten \finalonly\
models (Figure~\ref{fig:transport}). These answers fade soon after they are
computed, in swapped and unmodified runs alike.
\item \textbf{Answers become readable before their states become reusable
(Section~\ref{sec:temporal}).} Freezing each position $k$ from the loop at
which its answer first becomes readable, as per-position halting would and
with no other run or label involved, reduces endpoint accuracy from 100\% to
4.1\% under \sampledprefix\ and to 6.3\% under \matchedexit, two objectives
that retain answers well (Figure~\ref{fig:freeze}). Freezing two to four
loops later restores it. A released recurrent maze network that never loses
a solved maze solves none of 32 mazes when each pixel is frozen at its
answer's arrival, and all of them when frozen four iterations later.
\end{itemize}

These interventions separate the three properties; they do not identify a
microscopic cause of ordinary decline or give a general repair. The main
evidence comes from one synthetic task family and one architecture, with a
limited external check (Section~\ref{sec:discussion}).

\section{A controlled setting for computation and retention}
\label{sec:setting}

\begin{figure}[t]
 \centering
 \includegraphics[width=\linewidth]{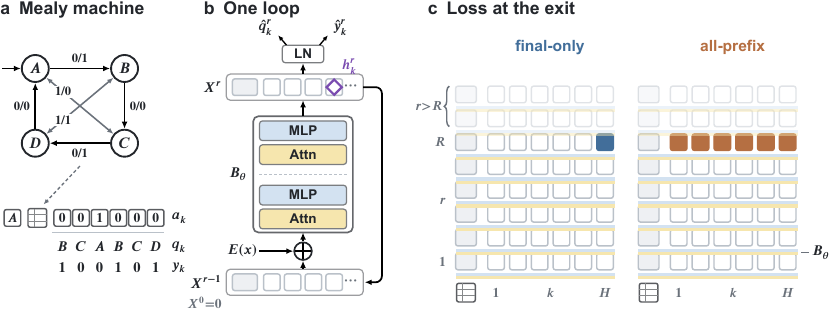}
 \caption{\textbf{Task, recurrence, and where the loss is applied.}
 (a) A Mealy machine (edge label: input/output), its in-context encoding,
 and the simulator targets $q_k$, $y_k$. (b) One loop of the shared block
 $B_\theta$; $\diamond$ marks where interventions replace $h_k^r$.
 (c) Each cell is one vector $h_k^r$ (position $k$, loop $r$). Colored cells
 receive the loss at the exit $R$; rows above $R$ are never trained and are
 where retention is measured.}
 \label{fig:setting}
\end{figure}

\paragraph{Task and model.}
A deterministic Mealy machine has a finite state set $Q$, input alphabet
$\Sigma$, output alphabet $\Gamma$, initial state $q_0$, transition function
$\delta:Q\times\Sigma\to Q$, and output function $\lambda:Q\times\Sigma\to\Gamma$.
Reading a sequence of input symbols, which we call actions,
$a_1,\ldots,a_H\in\Sigma$ ($H$ is the horizon), it computes
\begin{equation}
 q_k=\delta(q_{k-1},a_k),\qquad
 y_k=\lambda(q_{k-1},a_k),\qquad k=1,\ldots,H,
 \label{eq:mealy}
\end{equation}
so $q_k$ is the state after action $k$ and $y_k$ the output emitted on that
transition. In Figure~\ref{fig:setting}a, from $q_0=A$ the action $a_1=0$
follows the edge $0/1$, giving $q_1=B$ and $y_1=1$. An example $x$ lists
$q_0$, one token per table entry $(q,a)\mapsto(\delta(q,a),\lambda(q,a))$,
and the $H$ actions. At the token of action $k$ the model predicts $q_k$ and
$y_k$; these are targets, never inputs. The generator samples reachable,
behaviorally minimal machines with 4, 6, or 8 states, 2 or 4 input symbols,
and 2 output symbols;
randomized symbol names and row order discourage fixed-label shortcuts but do
not establish structural disjointness of sampled machines
(Appendix~\ref{app:methods}).

The model repeats a two-layer transformer block with width 128, four attention
heads, and feed-forward width 512. With input embeddings $E(x)$, it computes
\begin{equation}
 X^0=0,\qquad X^r=B_\theta\bigl(X^{r-1}+E(x)\bigr).
 \label{eq:model}
\end{equation}
Here $r$ indexes loops and $X^r$ collects the hidden vectors of all tokens
after loop $r$; training reads out after loop $R$, the exit.
The block's two layers have different parameters; the entire block is shared
across loops (Figure~\ref{fig:setting}b; details in Figure~\ref{fig:arch-detail}).
There are no positional or loop-depth embeddings; the causal attention mask
provides the architectural signal of action order.

The prefix (initial-state token and transition table) attends within itself, and action position $k$
attends to this prefix and actions $1{:}k$. Let $h_k^r$ denote its post-block
hidden vector. Shared state and output heads are trained jointly with the
model and read $\operatorname{LN}(h_k^r)$ after each loop. Hidden-vector
interventions replace $h_k^r$ before this final normalization and affect
subsequent loops.

\paragraph{Training objectives.}
Our main comparison fixes the training exit $R=H$, architecture, curriculum,
optimizer, and training duration, and changes the supervised positions at the
exit. Each batch shares one horizon $H$. With $\ell_{i,k}^{R}$ the sum of
state and output cross-entropies for example $i$, the batch objective is
\begin{equation}
 \mathcal L_S=\frac{1}{\sum_i |S_i|}
       \sum_i\sum_{k\in S_i}\ell_{i,k}^{R},\qquad
 S_i=\begin{cases}
       \{H\},&\finalonly,\\
       \{1,\ldots,H\},&\allprefix,\\
       \{K_i\},\ K_i\sim U\{1,\ldots,H\},&\sampledprefix.
      \end{cases}
 \label{eq:loss}
\end{equation}
Relative to \finalonly, \allprefix\ broadens position coverage and increases
the number of supervised state--output pairs per example from one to $H$.
\sampledprefix\ separates coverage from dense per-example supervision by
selecting one position independently and uniformly per example. At fixed
parameters and batch, its expected loss equals the mean \allprefix\ loss.
All three objectives apply these losses only at the exit, not at every loop
(Figure~\ref{fig:setting}c), and average over selected positions rather than
multiplying the loss by $H$.

The main training comparisons use 132,000 AdamW steps, batch size 256, and
the same 35-stage curriculum reaching $H=32$. After the initial $H=1$ stage,
each stage draws its maximum horizon with probability .5 and otherwise a
shorter horizon uniformly from 1 to one below that maximum.
Under \finalonly, every horizon $H\le 32$ therefore receives endpoint
supervision at loop $H$ as the curriculum progresses.
Because future actions cannot affect earlier positions, a length-$k$ input
and any extension of it have identical computations at positions $1{:}k$
when the machine presentation is unchanged. Shorter-horizon training thus
directly constrains position $k$ at loop $k$ even when that prefix appears in
a longer input. \allprefix\ and \sampledprefix\ additionally supervise
that position at exits $H>k$, requiring its answer to be readable $H-k$
loops beyond the diagonal $r=k$. The relation of this diagonal to the
measured first-correct frontier is evaluated in Section~\ref{sec:retention}.

Section~\ref{sec:retention} also varies the exit distribution (Slack-4 versus
Jitter-8) and tests gradient truncation by comparing progressive training
with \matchedexit\ at matched supervised exits.

\paragraph{Evaluation cohorts.}
The \emph{common cohort} compares \finalonly, \allprefix, \sampledprefix,
\progressive, and \matchedexit\ on model seeds 9--11 and one shared
evaluation stream; the existing \finalonly/\allprefix\ checkpoints are
re-evaluated. The \emph{original cohort} uses seeds 9--14 for the historical
training comparisons, including Slack-4/Jitter-8, and supplies six pairs of
\finalonly/\allprefix\ models for the position-16 interventions
(Section~\ref{sec:temporal}; Appendices~\ref{app:delayed} and \ref{app:receiver}). Its behavioral evaluations use a separate
stream. The \emph{position-8 transport audit} in Section~\ref{sec:transport}
uses ten \finalonly\ models, seeds 9--18. Overlapping checkpoints are not
independent model replications. Additional freezing, larger-model, and maze
cohorts are specified with their experiments in Section~\ref{sec:temporal}
and the appendices. No seed is selected for intervention success;
Appendices~\ref{app:methods} and \ref{app:training-controls} give the seeds,
streams, and pairing protocols.

\paragraph{Readouts and scoring.}
For one model and $N$ evaluation examples of length $H$, endpoint accuracy
at readout loop $r$ is the fraction for which both predictions at the last
action are correct:
\begin{equation}
 A(H,r)=\frac1N\sum_{i=1}^N
 \indicator\!\left[\hat q_{i,H}^{r}=q_{i,H}\;\land\;
                    \hat y_{i,H}^{r}=y_{i,H}\right].
 \label{eq:endpoint}
\end{equation}
\emph{Whole-trajectory exactness} requires both predictions at all $H$
positions of an example to be correct at the same loop $r$, and reports the
fraction of examples satisfying this conjunction. The main endpoint and
trajectory evaluations use 512 traces per model and length: 16 machines
with 32 traces each. Scores are computed within each model and then averaged
across models with equal weight (Appendix~\ref{app:methods}).

On a clean loop--position grid, $\tau(k)$ is the first evaluated loop at which
\emph{both marginal} state and output accuracies at position $k$ reach .99,
for the specified model, horizon, and evaluation set. This population
criterion differs from the example-level conjunctions above. We reserve
$\tau(k)$ for this measured quantity and write $s(k)$ for a specified readout
or capture schedule. The matched-frontier audit in Section~\ref{sec:transport}
uses the diagonal schedule $s(k)=k$; a common-exit readout uses $s(k)=r$ for
all positions. Fixed freezing schedules and their agreement with the clean
frontier are assessed separately in Section~\ref{sec:temporal}.
Section~\ref{sec:transport} introduces source--base pairs, their
counterfactual targets, and the corresponding suffix-exactness scores.

\section{Training to retain changes inference-budget robustness}
\label{sec:retention}

\begin{figure}[t]
 \centering
 \includegraphics[width=\linewidth]{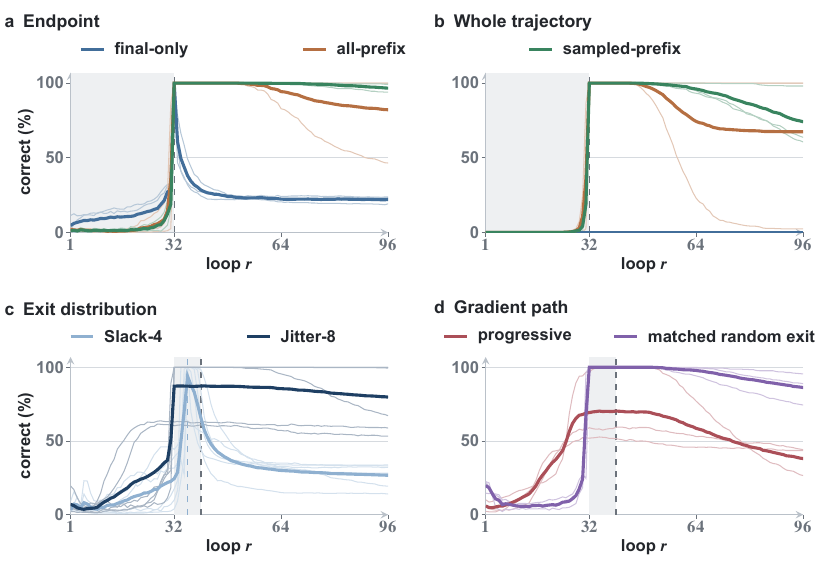}
 \caption{\textbf{Position coverage, exit distribution, and gradient propagation.}
 Endpoint joint accuracy (a,c,d) and whole-trajectory exactness (b: all 32
 state--output pairs correct) on 512 length-32 examples per model; thin curves
 are models, including incomplete learners, and thick curves their means.
 (a,b) Common cohort: one uniformly sampled supervised position per example
 preserves late accuracy without adding supervised positions. (c) Original
 cohort: Slack-4 and Jitter-8 match expected loop budgets. (d) Common cohort:
 progressive and matched random exit share supervised exits and forward work
 but differ in gradient truncation and backward work. Shading marks training
 exit support; the dashed line in (c) is Slack-4's exit. Neither comparison
 matches measured training FLOPs.}
 \label{fig:behavior}
\end{figure}

\paragraph{A correct endpoint need not survive more loops.}
Figure~\ref{fig:behavior}a,b evaluates the common cohort, holding each input
and checkpoint fixed across loops. \finalonly\ endpoint joint
accuracy falls from 100\% at loop 32 to 60.81\% at loop 33 and 22.33\% at
loop 64. Whole-trajectory exactness is 0\% at both loops 32 and 64.
These are losses on solved inputs of unchanged length. Original-cohort
results remain separate (Appendix~\ref{app:endpoint}).

\paragraph{One supervised position per example is enough for strong retention.}
At the same training exit $R=H$, \allprefix\ raises loop-64 endpoint accuracy
to 94.27\% on the common stream, but supplies more supervised positions.
\sampledprefix\ controls this difference: each example contributes one
uniformly chosen position's state and output losses. All three models attain
100\% endpoint and whole-trajectory accuracy at loop 32. At loops 64 and 96,
endpoint means are 99.80\% and 96.61\%, and whole-trajectory means are
96.03\% and 74.09\%. The corresponding \allprefix\ trajectory means are
74.28\% and 67.45\%.
Dense per-example supervision is therefore not necessary for the observed
retention gain. This three-seed result does not establish that sampling is
generally superior to \allprefix, nor remove the need for intermediate labels.

For both baselines trained at $R=H$, the original readability grids give a
first-correct frontier $\tau(k)=k$ (Appendix~\ref{app:trajectory});
Section~\ref{sec:transport} tests its causal role.

\paragraph{Varying the exit helps at matched expected loop budgets.}
The original-cohort comparison in Figure~\ref{fig:behavior}c retains
\finalonly\ supervision and compares Slack-4, $R=H+4$, with Jitter-8,
$R=H+U\{0,\ldots,8\}$. Both have conditional mean $H+4$; cumulative
batched block applications differ by only 0.062\%. At H32/loop64, endpoint
means are 30.01\% and 85.55\%, with Jitter-8 higher in every pair; at
loop96 they are 26.79\% and 79.79\%. Both readouts exceed Jitter-8's H32
training support of 32--40. The effect has tradeoffs: at the common loop36,
Slack-4 gives 93.98\% versus 87.27\%, and some seeds do not fully learn the
training length range. Appendix~\ref{app:budget} reports all seeds and work
accounting.

\paragraph{Truncated gradients do not help retention.}
\citet{bansal2022endtoend} prevent overthinking with progressive training,
which changes two things at once: the loss is applied at a random depth, and
gradients do not reach the loops before a random starting point. We adapt it
to this architecture and separate the two with a matched control. At each
update, both \progressive\ and \matchedexit\ add to a full branch ending at
$m=H+8$ a second branch with a random exit $n+k\geq H$. \progressive\
detaches this branch after its first $n$ loops; \matchedexit\
back-propagates through all $n+k$. Examples, exits, loss weights, and
updates are identical, so only the gradient path, and with it the backward
work, differs (Appendix~\ref{app:training-controls}).

In this recipe truncation impedes learning. Two of three \progressive\
models never exceed 60\% H32 endpoint accuracy, so their low late scores are
not forgetting; the third reaches 100\% at loops 40--48 and falls to 26.6\%
by loop 96. All three \matchedexit\ models reach 100\% at loop 32 and keep
97.9\% at loop 64 (Figure~\ref{fig:behavior}d). Like Jitter-8, this
random-exit objective keeps endpoint answers well past its training exits.
Its losses, like those of \finalonly, use only the last position, and it
keeps far fewer complete trajectories: 29.6\% are exact at loop 64, against
96.0\% under \sampledprefix. This is an FSM adaptation, not a reproduction
of the original architecture or a ranking of progressive methods.

\paragraph{Retention is a distinct, finite-window property.}
Supervision support does not fully explain learning or retention, and H32
retention does not guarantee improved length extrapolation.
On the common H64/loop96 evaluation, \sampledprefix\ has 82.68\% endpoint
accuracy versus 93.03\% for \matchedexit. The original length comparison also
has mixed outcomes (Appendix~\ref{app:length}).

\section{Computation is exact, but its answers are transient}
\label{sec:transport}

Section~\ref{sec:retention} found that each position's answer first becomes
readable at loop $k$, even in \finalonly\ models that get almost no complete
trajectory right at the trained exit. Readable does not mean used: a readout
can decode information that the computation ignores. We therefore test
whether the frontier state is what the rest of the execution runs on. If it
is, replacing it with another input's state should make the run follow that
input's execution exactly, and a simulator tells us what that execution is.

\begin{figure}[t]
 \centering
 \includegraphics[width=\linewidth]{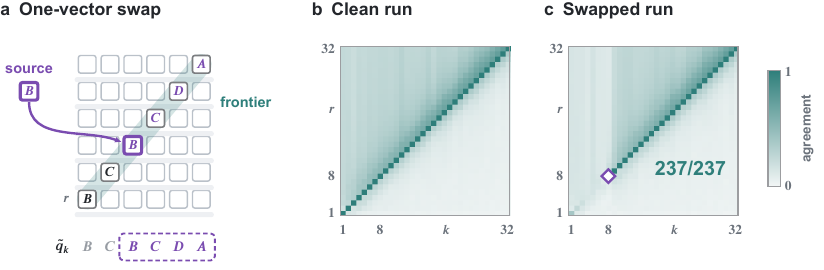}
 \caption{\textbf{One swapped state drives the exact remaining execution.}
 (a) Schematic, $H=R=6$, $c=3$. The source's vector at position 3 and loop 3,
 whose state is $B$, replaces the base run's; read at loop $k$ (the
 frontier), positions 4--6 then give $C$, $D$, $A$, the source's continuation
 $\tilde q_k$. (b,c) Ten \finalonly\ models, $H=32$, swap at position 8 and
 loop 8 ($\diamond$ in c). Shading is the fraction of pairs whose state and
 output at $(k,r)$ both match the target: the base run's own trajectory (b)
 or the source's continuation after the swap (c); means over 256 pairs and
 ten models. The bright diagonal is the frontier; the fading above it shows
 answers decaying after they are computed, and some agreement away from it
 is chance. 237/237: eligible pairs whose whole suffix ($k=9,\ldots,32$)
 matches along the frontier.}
 \label{fig:transport}
\end{figure}

\paragraph{Swapping one state between two inputs.}
We pair two inputs that share the machine, symbol assignment, and
transition-table presentation, and differ only in their actions up to
position $c$; their states $q_c^S$ and $q_c^B$ at $c$ differ
(Figure~\ref{fig:transport}a). Following interchange interventions
\citep{geiger2024finding}, we copy the full 128-dimensional hidden vector of
the \emph{source run} at position $c$ and loop $r$ into the \emph{base run},
\begin{equation}
 h_c^r\leftarrow h_{c,S}^r,
 \label{eq:swap}
\end{equation}
leave every other base position unchanged, and let the trained model
continue. No symbolic state or answer enters the network. Because the
remaining actions are shared, a base run that now carries the source's state
should continue exactly as the source run does. Running
Equation~\ref{eq:mealy} from $q_c^S$ on these actions gives this
counterfactual target $\tilde q_k$ for every $k>c$ \citep{geiger2021causal}.
A pair is \emph{suffix-exact} under a readout schedule $s(k)$ when every
state and output after $c$ matches the target. We score only \emph{eligible}
pairs, whose counterfactual and factual suffixes differ, so that a run
ignoring the swap cannot succeed (Appendix~\ref{app:methods}).

\paragraph{One swapped state drives the rest of the execution.}
In ten \finalonly\ models with $H=32$, we swap at position 8 and loop 8,
the loop at which position 8's answer first becomes correct in every model
($\tau(8)=8$; Appendix~\ref{app:methods}). Of 256 pairs, 237 are eligible. In every
model, all 237 follow the source exactly: every state and output at
positions 9 to 32, each read at loop $k$ (Figure~\ref{fig:transport}c). One
vector thus steers 24 further steps of exact symbolic execution. Copying a
vector back into its own run changes no prediction, and patch scans place the
damaging site on the same frontier (Appendix~\ref{app:methods}). The swap
shows that this state suffices to drive the computation; it does not show
that the state is a unique symbolic code.

\paragraph{The answers do not stay.}
In both the swapped and the clean run, agreement is highest on the diagonal
and fades above it (Figure~\ref{fig:transport}b,c): an answer is readable at
the loop that computes it, and answers computed early have faded by loop 32.
The near-zero whole-trajectory exactness of \finalonly\ models at the trained
exit (Section~\ref{sec:retention}) is therefore not a failure to compute: the
intermediate answers are computed exactly and then not kept.

\section{Answers become readable before their states become reusable}
\label{sec:temporal}

\begin{figure}[t]
 \centering
 \includegraphics[width=\linewidth]{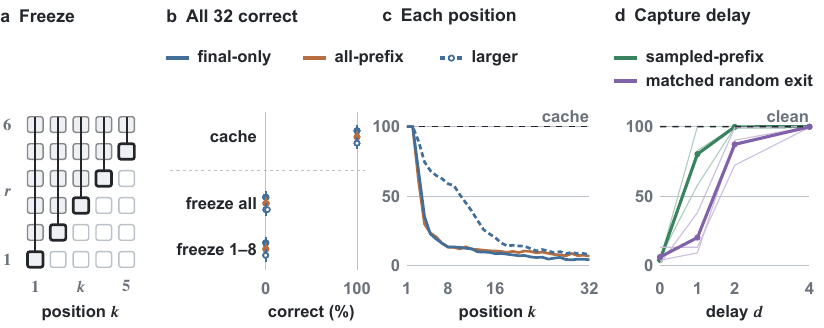}
 \caption{\textbf{Answers can be cached when they appear; their states can be
 frozen only a few loops later.}
 (a) Freezing captures each position's own hidden vector at loop $k$ and
 writes it back after every later block; other positions update normally,
 and no other run or label is involved. (b,c) Six \finalonly, three
 \allprefix, and three larger \finalonly\ models, 512 H32 examples each.
 (b) Fraction of examples with all 32 state--output pairs correct at loop 32.
 \emph{Cache} keeps each position's loop-$k$ prediction outside the model;
 \emph{freeze} holds the hidden vectors instead, for all positions or only
 positions 1--8. Dots are models; short marks are group means. (c) Accuracy
 of each position at loop 32 after freezing all positions (group means);
 the dashed line is the cache. (d) \sampledprefix\ and \matchedexit, three
 seeds each (thin lines) and their means (thick), with capture delayed by $d$
 loops: endpoint accuracy at the actual capture loop $32+d$, unchanged
 through loop 64; the dashed line is clean. Clean readouts at $s(k)=k$ are
 not perfectly correct for two \matchedexit\ seeds
 (Appendix~\ref{app:freeze-controls}).}
 \label{fig:freeze}
\end{figure}

\paragraph{Halting treats a solved position as finished.}
Section~\ref{sec:retention} showed that training can keep an answer readable
long after it is computed, and Section~\ref{sec:transport} that a single
frontier state drives the computation that follows. A position whose answer
has appeared and stays readable therefore looks finished, and adaptive
computation treats it that way. The Universal Transformer halts positions
individually and copies a halted position's state unchanged to every later
step \citep{dehghani2019universal}. Huginn stops iterating on a token once
successive steps give nearly the same output distribution, and later tokens
attend to its last computed key--value state \citep{geiping2025scaling}.
\citet{bansal2022endtoend} describe their networks, trained to avoid
overthinking, as holding on to the representation of a solution once it is
reached. These methods differ in when a position counts as finished. We test
the earliest point, the loop at which its answer first becomes readable. Our
setting makes that loop known: position $k$ first becomes correct at loop
$k$, and every position has an exact target.

\paragraph{Freezing applies this halting at the first readable loop.}
For each position $k$, we capture its hidden vector at loop $k$ and write it
back after every later block, as copying a halted state would. All other
positions update normally, frozen positions stay visible to attention, and no
other run or label is involved (Figure~\ref{fig:freeze}a). As a control,
\emph{readout caching} stores each position's loop-$k$ prediction outside
the model and leaves every hidden state untouched; the two differ only in
whether the answer or its state is kept. We call a state \emph{reusable} if
freezing it in this way leaves the later positions correct. We freeze six
\finalonly, three \allprefix, and three larger \finalonly\ models ($d=512$
and four layers per block, about 32 times the block parameters), and the
three \sampledprefix\ and three \matchedexit\ models of
Section~\ref{sec:retention}.

\paragraph{The answers are correct, but their states cannot yet be frozen.}
In the twelve models of Figure~\ref{fig:freeze}b, caching keeps all 32
answers correct, so every answer is correct when it is captured. Freezing the
states instead leaves at most 2\% of examples fully correct. Good retention
does not prevent this. \sampledprefix\ keeps 99.8\% of its endpoint answers
at loop 64 (Section~\ref{sec:retention}) and its cached answers are all
correct, yet freezing drops its endpoint accuracy from 100\% to 4.1\%.
\allprefix\ models, which keep every position readable at the exit, and
\matchedexit\ models collapse in the same way (Appendix~\ref{app:freeze}).
The answers stay readable; their states cannot yet stop changing.

\paragraph{Frozen states fail as inputs to later computation.}
The errors start just downstream of the frozen states
(Figure~\ref{fig:freeze}c). Positions 1 and 2 stay correct in every model,
and position 3 is the first to fail. At loop 3 it reads the loop-2 vectors
of positions 1 to 3, which differ from the clean run in one place only:
position 1 has missed its loop-2 update, which came after its answer was
already readable. That one missing update lowers the accuracy of position 3
to 68\% in the \finalonly\ models, and errors then compound along the
sequence. Freezing only positions 1--8 leaves positions
9--32 free to update, yet in the \finalonly\ models the endpoint is still
correct on at most 11.9\% of examples (Appendix~\ref{app:freeze}). A frozen
state still yields its own answer, as positions 1 and 2 show, but it no
longer works as input: later positions need the updates that upstream states
receive after their answers appear.

\paragraph{A state becomes reusable two to four loops after its answer.}
Delaying the capture by $d$ loops shows when a state can be frozen
(Figure~\ref{fig:freeze}d). With a two-loop delay, \sampledprefix\ recovers
to 99.8\% endpoint accuracy; \matchedexit\ needs four loops to reach the same
level (Appendix~\ref{app:freeze-controls}). An answer is readable at loop $k$,
but the state that carries it becomes reusable only a few loops later, and
how many loops depends on the model.

\paragraph{The maze network trained against overthinking behaves the same way.}
We apply the same test to the released maze network of
\citet{bansal2022endtoend}, the model described above as holding on to its
solution. Trained with 30 iterations, it solves all 32 test mazes of size
$9\times9$ at every iteration from 25 to 120. Since mazes have no
per-position frontier, we freeze each pixel at its answer's arrival, the last
iteration at which its clean prediction changes. No maze is then solved;
freezing two or four iterations later solves 28 and 32, and caching the
answers solves all 32 (Appendix~\ref{app:freeze}; one checkpoint, 32 mazes
per size). The network does hold on to its solution, but only a few
iterations after each answer arrives.

\paragraph{What freezing shows.}
An answer readable at loop $k$ is thus not yet carried by a reusable state,
even in models that retain answers almost perfectly; the state becomes
reusable two to four loops later. Retention and reuse are different
properties. For per-position halting in such models, stopping a position when
its answer first becomes readable is too early: its state must keep updating
for a few loops, or only its answer should be kept. The experiments do not show
that information is lost, since cached answers stay correct, and they do not
identify a mechanism. The model has no positional or loop embeddings, so the
evolving state may also carry timing or order information that later
positions use to find their predecessors, and freezing would hold that
information as well. Our tests do not separate these possibilities; under
either, a state cannot be frozen when its answer first appears. A source
vector transplanted from another run shows the same dependence on timing
(Appendix~\ref{app:receiver}).

\section{Related work}

\paragraph{Extra iterations in recurrent reasoners.}
Recurrent networks trained on small instances can solve larger ones by
iterating longer \citep{schwarzschild2021learn}. \citet{bansal2022endtoend}
show that excessive iteration can degrade answers (``overthinking''), and
prevent this with input recall and progressive training from detached
intermediate states. They also test recovery after feature-map interventions
and track feature convergence. We adapt their progressive loss in
Section~\ref{sec:retention} and freeze their released maze network in
Section~\ref{sec:temporal}. \citet{anil2022path} relate additional
computation to path independence: convergence to the same steady state
regardless of initialization. We do not claim novelty for overthinking,
variable exits, or mid-computation substitution. What is new is the
measurement: exact symbolic targets for every position, loop, and
counterfactual, which let us test a state at the loop where its answer
first appears.

These works assess stability through outputs, feature maps, or final
convergence; we ask whether a state can be held from the moment its clean
readout becomes correct. This is a position-level counterpart to path
independence, not a test of it: we do not vary initialization.

\paragraph{Recurrent computation and training requirements.}
Universal and looped transformers connect shared parameters to adaptive
computation, programmable execution, and length generalization
\citep{dehghani2019universal,giannou2023looped,fan2025looped}; recurrent-depth
language models extend this motivation to latent test-time compute
\citep{geiping2025scaling}. \citet{zhang2026recurrence} study budget-dependent
frontiers, causal damage cones, and frontier-state transplants. We build on
these observations rather than claiming novelty for frontiers or transport.
A first-correct frontier differs from a convergence time that requires an
answer to remain correct.

Stochastic stopping can stabilize budget-dependent behavior without better
extrapolation \citep{kuo2026stabilizing}; adaptive-depth studies distinguish
trajectory formation from readout selection \citep{popescu2026adaptive}.
Our comparisons instead change supervised positions at a fixed exit, or exit
variability at matched expected loops; sampling one position estimates the
all-position loss rather than introducing a new loss principle.

\paragraph{Readable and causally usable representations.}
\citet{wang2026jacobian} distinguish readout, transport, and steering across
recurrent depth in Ouro and Huginn, and associate verbalizing injected content
with per-iteration supervision. Our \allprefix\ loss applies only at the
exit; we add paired training comparisons, exact symbolic targets, and local
downstream restoration that separates downstream state from execution phase.

State-tracking studies identify alternative transformer algorithms and effects
of intermediate objectives \citep{liu2023transformers,zhang2025finite,li2025how},
and interchange interventions test such algorithmic hypotheses
\citep{geiger2021causal,geiger2024finding}. Our full-vector intervention
establishes sufficiency in a specified machine context, not a unique minimal
encoding or complete algorithm identification.

\section{Discussion and limitations}
\label{sec:discussion}

Computing an answer, retaining its readout, and reusing its representation
are different properties, and each needs its own test: correctness at one exit
does not test retention, and retained predictions do not test whether a hidden
state can be held or reused for continuation. Supervised positions, exit distributions, and
gradient propagation all change retention, and poor late accuracy can conceal
a coherent earlier counterfactual computation. Retained answers need not
reside in reusable states: freezing a position when its answer first becomes
readable disrupts later computation, including under sampled-position
training, and the state becomes reusable only a few loops later; a
transplanted source likewise needs the downstream position reset before it
takes effect. We have not tested downstream resets on the new training controls or
stochastic-budget models, so their retention gains do not establish the same
downstream mechanism.

These full-vector, post-block interventions do not identify a unique
encoding, microscopic cause of forgetting, or general repair; they can change
nonsymbolic features or restart computation, and downstream recovery concerns
one step at one site and age. The scope is one synthetic execution family and one principal
architecture, plus a limited external maze check. Original training
comparisons use six paired seeds; new controls use three. Re-evaluating
overlapping checkpoints is not independent model replication. Sampled
supervision still needs intermediate labels. Budget controls match expected
loops, and progressive controls match forward work but differ in backward
work; neither matches measured total FLOPs. Finite-window retention does
not establish hidden-state convergence or uniform length extrapolation.

\subsection*{AI use statement}
The core idea of this work is the authors' own. Generative AI tools assisted with code
implementation, experiment and control iteration, literature search, translation, manuscript
editing, and figure code. The authors verified the analyses, citations, and final artifacts and
take full responsibility for the work.

\subsection*{Ethics statement}
Machines and traces are procedurally generated; the work uses no personal or
human-subject data. The objectives studied here do not ensure the reliability
of deployed systems.

\subsection*{Reproducibility statement}
Appendix~\ref{app:methods} specifies machine generation, the network,
optimization, curriculum, and scoring. Appendices~\ref{app:results}--\ref{app:natural}
give every training and intervention protocol with per-seed results, and
Appendix~\ref{app:provenance} records the checkpoints, evaluation streams,
pair identities, and hashes behind the principal quantities.

\bibliography{references}
\bibliographystyle{iclr2027_conference}
\clearpage
\appendix
\setcounter{figure}{0}
\setcounter{table}{0}
\renewcommand{\thefigure}{A\arabic{figure}}
\renewcommand{\thetable}{A\arabic{table}}
\renewcommand{\theHfigure}{appendix.\arabic{figure}}
\renewcommand{\theHtable}{appendix.\arabic{table}}
\section{Experimental methods and scoring}
\label{app:methods}

\subsection{Machine generation and input encoding}
At each stage, the data generator samples the state count and input-alphabet
size independently and uniformly from the active lists in
Table~\ref{tab:curriculum}; the output alphabet has two symbols. Generation
fixes an internal initial state $S_0$ and first makes every state reachable.
In a random order of the other states, each is attached through a uniformly
sampled unused state--input cell of an already reachable state. Remaining
transition destinations are sampled uniformly from $Q$, and each cell's
output is sampled uniformly from $\Gamma$. Candidates are resampled until
behaviorally minimal and covering both output symbols. These construction
and acceptance rules make the accepted transition-table entries dependent.

Each sampled machine contributes up to 32 action traces, all starting at
$S_0$, with actions drawn independently and uniformly from its input alphabet.
A training batch contains 256 examples of the same sampled horizon. The
online stream, including machine generation, action traces, and presentation
randomization, is reproducible from its generator state.

The initial state is one token. Every transition is represented by one
factorized token containing its source state, input symbol, destination state,
and output symbol. Action tokens follow the table and contain only their
input symbol. Token embeddings sum a kind embedding and the embeddings of the
present symbolic fields. The global namespaces contain 16 state IDs, 8 input
IDs, and 4 output IDs, although an individual machine uses only 4/6/8 states,
2/4 inputs, and 2 outputs. Each presentation assigns local symbols to randomly
selected namespace IDs and shuffles the transition rows. Training does not
force traces from one machine to share the same presentation; source--base pairs
explicitly share a presentation to control that variable.

The attention mask lets initial-state and table tokens attend within the
prefix, and action $k$ attend to the prefix and actions up to and including
$k$. It blocks all access to future actions. Padded positions are masked and
zeroed through the computation. A causal action prefix therefore has the same
unmodified computation when followed by additional actions of the same
presented machine. This structural property helps explain why endpoint
training at shorter lengths constrains earlier positions in longer examples;
it does not prove a particular internal algorithm.

Training uses data seed 20260901. The original six-seed budget comparisons use
evaluation seed 20260909; the five-condition, three-seed comparison uses
20260923, with the old baselines re-evaluated on that same stream. Both use
four batches of 128 examples per model and length: 16 sampled machines with
32 traces each, not 512 independent machines. The new six-model freeze
follow-up uses stream 20260924. Actual generator seeds add $1009H$ to the
specified stream. The original training logs use evaluation seed 20260902.
These are separate evaluation streams. We do not
combine their metrics as if they came from one paired sample set. Randomized
presentations and independently drawn evaluation streams do not guarantee
that every machine graph is structurally absent from training.

\subsection{Network, optimization, and curriculum}
The shared block contains two pre-normalized transformer layers, each with
four attention heads, width 128, a width-512 feed-forward network, and residual
connections. Dropout is zero. The recurrence begins with a zero hidden matrix
and adds the same input embedding before every block application. There is no
learned recurrent gate, loop embedding, positional embedding, or enabled
layer scaling. State and output linear heads follow a shared final layer
normalization. The heads have 16 and 4 classes, respectively. All reported
hidden-vector replacements occur after the shared block and before this
final normalization, exactly as in Equation~\ref{eq:swap}.

\begin{figure}[t]
 \centering
 \includegraphics[width=\linewidth]{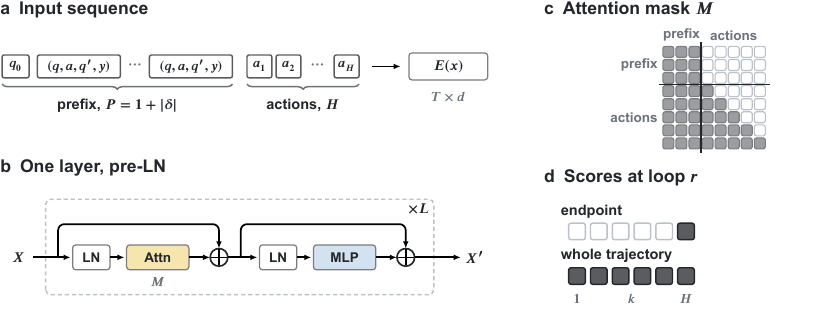}
 \caption{\textbf{Network details.} (a) An example is one sequence: the
 initial state $q_0$, one factorized token $(q,a,q',y)$ per transition, and
 the $H$ action symbols. It is embedded once as $E(x)$, with $T=P+H$ rows of
 width $d$, and added before every application of $B_\theta$. (b) $B_\theta$
 stacks $L$ pre-normalized layers with distinct weights: $L=2$, $d=128$, four
 heads, and MLP width 512 in the main models; $L=4$ and $d=512$ in the larger
 models. There is no dropout, recurrent gate, layer scaling, or positional or
 loop embedding. (c) Rows are queries and columns keys; filled cells are
 visible. Prefix tokens attend within the prefix, and action position $k$
 attends to the prefix and positions $1{:}k$. Schematic with three prefix
 tokens and five actions; padding is masked. (d) At loop $r$, endpoint
 accuracy scores position $H$; whole-trajectory exactness requires all $H$
 positions of an example to be correct at that same loop. A position is
 correct when its state and output are both correct.}
 \label{fig:arch-detail}
\end{figure}

Figure~\ref{fig:arch-detail} summarizes the input sequence, one layer, the
attention mask, and the two per-loop scores.

AdamW uses initial learning rate $3\times10^{-4}$ with cosine decay to zero,
weight decay .01, gradient norm clipping at 1, and batch size 256. Both task
losses have weight 1 and use mean
cross-entropy over the selected valid positions. Training runs for the fixed
132,000-step curriculum in Table~\ref{tab:curriculum}; stage checks are audits,
not a rule for selecting a checkpoint or stopping a failing seed. Checkpoints
are saved at fixed steps, and all main comparisons use step 132,000.

\begin{table}[ht]
 \centering
 \caption{\textbf{The 35-stage curriculum.} A maximum horizon is sampled with
 probability .5 after the first stage; the remaining draws come from shorter
 horizons. All stages include the shorter-length range from 1. Rows grouping
 several stages retain their individual training allocations below.}
 \label{tab:curriculum}
 \small
 \begin{tabular}{lrrp{5.9cm}}
  \toprule
  Maximum horizon & Stages & Steps & Machine distribution\\
  \midrule
  1 & 1 & 2,000 & 4 states, 2 inputs, 2 outputs\\
  2 & 4 & 7,000 & 2,000 steps each with 4, then 4/6, then 4/6/8 states;
                              1,000 steps adding the 4-input cases\\
  3 & 1 & 3,000 & 4/6/8 states, 2/4 inputs, 2 outputs\\
  4 & 1 & 8,000 & Same full distribution\\
  5--32 & 28 & 112,000 & 4,000 steps per successive maximum horizon\\
  \bottomrule
 \end{tabular}
\end{table}

\subsection{Readouts, thresholds, and exactness}
Let $Q_{i,k}^r$ and $Y_{i,k}^r$ be indicators of correct state and output for
sample $i$, position $k$, and loop $r$. The endpoint curve reports
$N^{-1}\sum_i Q_{i,H}^r Y_{i,H}^r$. The grid criterion instead requires both
$N^{-1}\sum_i Q_{i,k}^r\geq .99$ and
$N^{-1}\sum_i Y_{i,k}^r\geq .99$. Passing this criterion is not the same as a
per-sample guarantee that every state and output is correct. The first loop
passing it on the clean grid defines $\tau(k)$ for that model, horizon, and
evaluation set. This measured quantity is distinct from a prescribed
readout or capture schedule $s(k)$. Hold includes consecutive passing loops
from $\tau(k)$, including the first. If the measured window ends while the
criterion still holds, the observed hold is a right-censored lower bound.

For the whole-trajectory metric in Figure~\ref{fig:behavior}, an example is
trajectory-exact at readout loop $r$ when every one of its $H$ state--output pairs
is correct at that same readout:
\[
 E_i^r = \prod_{k=1}^{H} Q_{i,k}^r Y_{i,k}^r,
 \qquad
 \operatorname{TrajExact}(r)=N^{-1}\sum_i E_i^r.
\]
This example-level conjunction is stricter than either the endpoint event or
the grid's two marginal population thresholds.

For a target sequence $(\bar q_{i,k},\bar y_{i,k})$ and readout schedule $s(k)$,
suffix exactness on an eligible set $\mathcal E$ is
\begin{equation}
 \operatorname{Exact}_{\mathcal E}(s)
 =\frac1{|\mathcal E|}\sum_{i\in\mathcal E}
   \prod_{k=c+1}^{H}
   \indicator[\hat q_{i,k}^{s(k)}=\bar q_{i,k}]
   \indicator[\hat y_{i,k}^{s(k)}=\bar y_{i,k}].
 \label{eq:suffix}
\end{equation}
The ordinary target is the base run's factual trajectory. The source target
is the simulated counterfactual. Changing this target changes the estimand;
changing $s(k)$ changes when it is observed. A wrong prediction that differs
from both trajectories is not successful transport.

\subsection{Position-8 transport audit}
The position-8 audit uses ten serial \finalonly\ models, seeds 9--18, with
$H=32$ and a single replacement at loop 8. On a separate clean grid
(evaluation stream 20260903, 512 examples per model), all ten models have
$\tau(k)=k$ at every position $k=1,\ldots,32$, with state and output
accuracy 1.000 on the diagonal. The same 256 different-state pairs
are reused across seeds. A pair is eligible for suffix scoring if at least
one state or output in positions 9--32 differs between its factual and source
targets; there are 237 such pairs. Conditional exactness uses that denominator,
not 256 or ten times 237 independent task samples. The matched-frontier
schedule is $s(k)=k$ and the common-exit schedule is $s(k)=32$.
Eligibility requires a difference somewhere in the suffix, not at every
suffix position. Factual and counterfactual states may later merge. Complete
suffix agreement consequently does not establish that their distinction
remains causally active at every later position.

At the common exit $s(k)=32$, no eligible suffix is exact in seven models and
one is exact in three. The clean base run is not jointly readable there
either (mean per-position agreement 38.1\% over positions 9--32), so this
contrast concerns readout time; it shows neither erased information nor a
unique state code.

\paragraph{Localization and intervention controls.}
Independent patch scans replace a width-one window of positions at loop
$l$. The maximal damage falls on position $l$, the observed loop--position
frontier, in 320/320 tested loop--seed cells. Prefix replacements that
overwrite positions before an anchor are predominantly safe until the
anchor passes this boundary: the first damaging anchor is $l+1$ in 309/310
testable cases; the exception (seed 15, loop 31) is retained. Identical
self-replacements do not change any predictions. Sources with the same
symbolic state but a different history can still cause damage; they are not
equivalent to self-replacement controls. Together, these results support a
localized, causally sufficient computation under the specified
intervention. They do not establish a pure or unique symbolic state code,
nor use independent of machine context or execution phase.

The saved clean readouts in this position-8 audit include per-position
accuracy, not clean whole-suffix exactness. In particular, a saved field for
the \emph{base-target exactness of a patched run} is not a clean
baseline. We use explicit unpatched suffix metrics from the separate
position-16 experiment when making clean-suffix comparisons.

\subsection{Paired delayed and phase-crossing cohorts}
The paired retention experiment freezes six models per supervision condition,
seeds 9--14. It samples 256 different-state pairs and 256 same-state pairs per
checkpoint, in sixteen batches of sixteen per pair kind. Pair identities are
matched across conditions and seeds. The site stays at position $c=16$; patch
age $a\in\{0,1,2,4,8,16\}$ means taking and inserting the source at loop
$16+a$, rather than holding a loop-16 vector and inserting it later.

Shifted-frontier scoring uses $s(k)=k+a$. Fixed-exit scoring uses $s(k)=32$ or
48. At fixed exit 32, increasing age reduces the remaining update count, and
age 16 leaves no post-patch update. Shifted-frontier scoring fixes the update
count to each target but still changes both source representation and base-run
context. Neither age curve by itself identifies information decay.

The 242 different-state pairs with distinct complete suffix targets are the
fixed denominator for source suffix exactness across all these schedules.
Ordinary clean suffix exactness uses all 256 pairs. For a single target at
distance $d=1,2,4,8,16$, factual and source trajectories can have merged;
single-target eligible counts are respectively 242, 207, 143, 90, and 35.
Rates for an empty eligible set are undefined and stored as null, not zero.

The subsequent compatibility experiment reuses the pair identities and
checkpoint configurations. It crosses source collection at $c$ or $c+a$ with
base insertion at those same two loops, reading position $c+d$ after exactly
$d$ further updates. The main descriptive slice is $a=4,d=1$, with 242
eligible pairs. Its extent comparison instead inserts at loop $p=c+a$ and
reads the unpatched position $p+d$ at loop $p+d$. This latter slice has 121
eligible pairs, and the old site is five positions from the target while the
current frontier is one position away. The two experiments are not interchangeable
scoring slices. All 384 batches across twelve checkpoint conditions were
validated for the compatibility cohort; the separate pilot is excluded.

\subsection{Controls and statistical interpretation}
True self-replacement copies the base run's own vector at the same site and
time, and must preserve predictions exactly. Same-symbolic-state sources come
from another history and are a substantive intervention, not an identity
control. A positive source score is interpreted alongside the clean base run's
agreement with its own target and with the source target. Chance coincidence
with a source target is not causal transfer. A clean factual score is a
reference level, not a formal upper bound on intervention performance.

Means first aggregate the specified pair/sample population within each model,
then average models with equal weight. Model seeds are independent parameter
initializations, whereas reused machine/pair samples are paired observations.
We preserve per-seed outcomes rather than converting task pairs into a larger
number of independent training runs. All intervals and differences in the
main text are descriptive; no null-hypothesis significance claim is based on
treating those repeated pairs as independent models.

\section{Additional results and their scope}
\label{app:results}

\subsection{Common-stream training controls}
\label{app:training-controls}
We compare five training conditions on the same three model seeds 9--11,
checkpoint step 132,000, and held-out stream 20260923. The existing
\finalonly{} and \allprefix{} checkpoints are re-evaluated; no baseline is
retrained or selected by its score. Each model receives four batches of 128
examples at each horizon. The generator seed is $20260923+1009H$, with
identical generator configuration and batch layout across conditions. The
H32 sweep covers $r=1,\ldots,96$ and the H64 sweep $r=1,\ldots,192$.
Pairing rests on this deterministic generator specification; the original
training-control evaluations did not save raw input hashes.
Model seeds, rather than examples or loop counts, are the replication unit.
The historical six-seed stream-20260909 results below remain a separate
cohort and are not pooled into these comparisons.

\paragraph{One supervised position per example.}
The new runs retain the original width 128, four attention heads, two shared
layers, additive input injection, batch size 256, AdamW optimizer, and
35-stage, 132,000-update curriculum. Stage progression and the cosine
learning-rate schedule are fixed, with no accuracy gate. Under
\sampledprefix{}, the training budget is $R=H$, as for the two serial
baselines. Each example independently selects one position uniformly from
$1,\ldots,H$ and applies the original state and output cross-entropies there.
If $\ell_{b,k}$ denotes their weighted sum, the mean loss satisfies
\[
 \mathbb{E}\!\left[\frac{1}{B}\sum_{b=1}^{B}\ell_{b,J_b}\right]
 =\frac{1}{BH}\sum_{b=1}^{B}\sum_{k=1}^{H}\ell_{b,k},
 \qquad J_b\sim U\{1,\ldots,H\}.
\]
Thus it is an unbiased estimator of the mean \allprefix{} loss. It changes
position coverage while retaining one supervised position per example;
it still requires intermediate trajectory labels and is not label-free.
The saved control protocol, rather than the legacy configuration field
\texttt{supervision=final\_only}, identifies this objective.

\paragraph{Progressive and matched random exits.}
We adapt the progressive loss of \citet{bansal2022endtoend} to the same FSM
Transformer. At horizon $H$, set $m=H+8$. Propose
$n\sim U\{0,\ldots,m-1\}$ and $k\mid n\sim U\{1,\ldots,m-n\}$;
reject and redraw the complete pair when $n+k<H$.
The accepted joint law is proportional to $[m(m-n)]^{-1}$ on
$H\leq n+k\leq m$, so neither accepted $n$ nor the exit is uniform.
Both conditions optimize $\tfrac12 L_m+\tfrac12 L_{n+k}$ using endpoint
labels. The $m$-loop branch has full gradients. For the random-exit branch,
\progressive{} runs $n$ loops without a gradient graph, detaches the hidden
state, and differentiates through the remaining $k$ loops. Input embeddings
are recomputed with gradients for injection into this continuation and
absolute loop indices are preserved. The $n=0$ case starts normally with
gradients. \matchedexit{} uses the same examples, draws, exits, and loss
weights, but differentiates through all $n+k$ loops. At fixed weights and
zero dropout, the two losses have equal values but different gradients.
This is an FSM adaptation, not a reproduction of the original convolutional
architecture, concatenated recall, or exit support; it is also distinct
from the uniform Jitter-8 contract below.

\paragraph{Compute accounting.}
Per update, both exit conditions execute $F=m+n+k$ forward blocks.
The numbers of differentiable blocks are $B_{\rm prog}=m+k$ and
$B_{\rm matched}=m+n+k$. Using a shape-aware matrix-multiply proxy with
forward block cost $bNL(12d^2+2Nd)$ and backward cost twice that amount
($b$ batch size, $N$ padded sequence length, $L$ shared layers), total work
is $5.919\times10^{16}$ MACs for \progressive{} and
$7.438\times10^{16}$ for \matchedexit{} per run. These exclude heads,
embeddings, normalization, optimizer, and evaluation; they are proxies,
not measured total FLOPs. Examples, updates, supervised exits, and forward
work are matched, but backward work is not. The removed gradient path is
the experimental intervention.

\begin{table}[t]
 \centering
 \small
 \setlength{\tabcolsep}{4pt}
 \caption{\textbf{Common-stream H32 training comparison.} Three-seed means
 in percent, all seeds 9--11, 512 examples per model, stream 20260923.
 $A_r$ is endpoint joint accuracy and $T_r$ is whole-trajectory exactness
 at loop $r$. The serial conditions train at $R=32$; the two exit controls
 supervise exits in 32--40. All rows use the same evaluated inputs.}
 \label{tab:control-h32}
\begin{tabular}{lrrrrrrr}
\toprule
Condition & $A_{32}$ & $A_{40}$ & $A_{64}$ & $A_{96}$ & $T_{32}$ & $T_{64}$ & $T_{96}$ \\
\midrule
\finalonly & 100.00 & 28.39 & 22.33 & 22.07 & 0.00 & 0.00 & 0.00 \\
\allprefix & 100.00 & 100.00 & 94.27 & 82.16 & 100.00 & 74.28 & 67.45 \\
\sampledprefix & 100.00 & 100.00 & 99.80 & 96.61 & 100.00 & 96.03 & 74.09 \\
\matchedexit & 100.00 & 100.00 & 97.85 & 86.26 & 74.09 & 29.56 & 11.52 \\
\progressive & 69.27 & 70.31 & 58.53 & 38.09 & 12.37 & 0.00 & 0.00 \\
\bottomrule
\end{tabular}

\end{table}

\begin{table}[t]
 \centering
 \scriptsize
 \setlength{\tabcolsep}{3.5pt}
 \caption{\textbf{Every model in the H32 training comparison.} Same metric,
 stream, and population as Table~\ref{tab:control-h32}; no seed is excluded.
 Peak $A$ is the maximum endpoint accuracy over the complete $r=1$--96
 sweep and is reported to distinguish failure to learn from later loss.}
 \label{tab:control-h32-seeds}
\begin{tabular}{lrrrrrrrrr}
\toprule
Condition & Seed & Peak $A$ & $A_{32}$ & $A_{40}$ & $A_{64}$ & $A_{96}$ & $T_{32}$ & $T_{64}$ & $T_{96}$ \\
\midrule
\finalonly & 9 & 100.00 & 100.00 & 27.15 & 24.02 & 23.24 & 0.00 & 0.00 & 0.00 \\
\finalonly & 10 & 100.00 & 100.00 & 33.59 & 20.51 & 19.14 & 0.00 & 0.00 & 0.00 \\
\finalonly & 11 & 100.00 & 100.00 & 24.41 & 22.46 & 23.83 & 0.00 & 0.00 & 0.00 \\
\midrule
\allprefix & 9 & 100.00 & 100.00 & 100.00 & 100.00 & 100.00 & 100.00 & 100.00 & 100.00 \\
\allprefix & 10 & 100.00 & 100.00 & 100.00 & 100.00 & 100.00 & 100.00 & 100.00 & 100.00 \\
\allprefix & 11 & 100.00 & 100.00 & 100.00 & 82.81 & 46.48 & 100.00 & 22.85 & 2.34 \\
\midrule
\sampledprefix & 9 & 100.00 & 100.00 & 100.00 & 100.00 & 99.22 & 100.00 & 100.00 & 98.05 \\
\sampledprefix & 10 & 100.00 & 100.00 & 100.00 & 100.00 & 96.68 & 100.00 & 94.92 & 60.55 \\
\sampledprefix & 11 & 100.00 & 100.00 & 100.00 & 99.41 & 93.95 & 100.00 & 93.16 & 63.67 \\
\midrule
\matchedexit & 9 & 100.00 & 100.00 & 100.00 & 100.00 & 95.51 & 99.41 & 75.39 & 32.23 \\
\matchedexit & 10 & 100.00 & 100.00 & 100.00 & 94.14 & 74.41 & 38.48 & 1.17 & 0.20 \\
\matchedexit & 11 & 100.00 & 100.00 & 100.00 & 99.41 & 88.87 & 84.38 & 12.11 & 2.15 \\
\midrule
\progressive & 9 & 59.38 & 58.98 & 59.38 & 52.93 & 44.14 & 0.78 & 0.00 & 0.00 \\
\progressive & 10 & 100.00 & 96.88 & 100.00 & 76.95 & 26.56 & 36.33 & 0.00 & 0.00 \\
\progressive & 11 & 52.73 & 51.95 & 51.56 & 45.70 & 43.55 & 0.00 & 0.00 & 0.00 \\
\bottomrule
\end{tabular}

\end{table}

\paragraph{Attainment and retention.}
The first four conditions reach 100\% endpoint accuracy at $r=32$ for all
three seeds. \sampledprefix{} retains 99.80\% endpoint accuracy and 96.03\%
whole-trajectory exactness at $r=64$. The comparison supports an effect of
supervision coverage without increasing the number of supervised positions
per example. The small seed cohort and point estimates do not establish
uniform or statistically demonstrated superiority over \allprefix{}.
In contrast, \progressive{} seeds 9 and 11 peak at only 59.38\% and
52.73\% endpoint accuracy: their low scores cannot be treated solely as
forgetting. Seed 10 reaches 100\% at $r=40$ and $48$, then falls to 76.95\%
at $r=64$ and 26.56\% at $r=96$. This fixed recipe provides no evidence
of an additional progressive advantage, but does not establish a general
ranking of progressive methods. The different outcomes under matched exits
also show that supervision support alone does not determine retention;
the gradient path matters in this comparison.

\begin{table}[t]
 \centering
 \small
 \setlength{\tabcolsep}{4pt}
 \caption{\textbf{H64 extrapolation of all five training conditions.}
 Every seed and its condition mean, 512 examples per model, stream 20260923.
 $A_r$ and $T_r$ are endpoint and whole-trajectory accuracy in percent.
 H64 exceeds the trained horizon range for every condition. These scores
 are length extrapolation, not H32 over-budget retention.}
 \label{tab:control-h64}
\begin{tabular}{lrrrrrrr}
\toprule
Condition & Seed & $A_{64}$ & $A_{72}$ & $A_{128}$ & $A_{192}$ & $T_{64}$ & $T_{128}$ \\
\midrule
\finalonly & 9 & 100.00 & 21.68 & 18.16 & 17.97 & 0.00 & 0.00 \\
\finalonly & 10 & 93.36 & 33.59 & 13.67 & 13.48 & 0.00 & 0.00 \\
\finalonly & 11 & 100.00 & 27.34 & 20.31 & 19.92 & 0.00 & 0.00 \\
\finalonly & Mean & 97.79 & 27.54 & 17.38 & 17.12 & 0.00 & 0.00 \\
\midrule
\allprefix & 9 & 100.00 & 100.00 & 99.61 & 96.29 & 100.00 & 97.85 \\
\allprefix & 10 & 100.00 & 100.00 & 100.00 & 99.22 & 100.00 & 100.00 \\
\allprefix & 11 & 30.08 & 27.73 & 24.80 & 22.66 & 0.20 & 0.00 \\
\allprefix & Mean & 76.69 & 75.91 & 74.80 & 72.72 & 66.73 & 65.95 \\
\midrule
\sampledprefix & 9 & 100.00 & 99.80 & 89.45 & 73.83 & 98.83 & 60.55 \\
\sampledprefix & 10 & 82.03 & 81.45 & 76.95 & 66.41 & 73.44 & 17.77 \\
\sampledprefix & 11 & 75.59 & 75.00 & 57.62 & 41.60 & 65.43 & 14.26 \\
\sampledprefix & Mean & 85.87 & 85.42 & 74.67 & 60.61 & 79.23 & 30.86 \\
\midrule
\matchedexit & 9 & 61.13 & 89.26 & 68.75 & 46.68 & 45.70 & 3.71 \\
\matchedexit & 10 & 99.61 & 100.00 & 72.66 & 48.83 & 1.37 & 0.00 \\
\matchedexit & 11 & 97.46 & 99.61 & 89.84 & 65.62 & 11.33 & 0.20 \\
\matchedexit & Mean & 86.07 & 96.29 & 77.08 & 53.71 & 19.47 & 1.30 \\
\midrule
\progressive & 9 & 44.92 & 43.55 & 39.26 & 31.84 & 0.00 & 0.00 \\
\progressive & 10 & 27.54 & 24.41 & 5.47 & 3.32 & 0.00 & 0.00 \\
\progressive & 11 & 29.30 & 29.69 & 19.53 & 16.41 & 0.00 & 0.00 \\
\progressive & Mean & 33.92 & 32.55 & 21.42 & 17.19 & 0.00 & 0.00 \\
\bottomrule
\end{tabular}

\end{table}

\paragraph{Length boundary.}
No condition improves all H64 readouts. At $r=64$, mean endpoint accuracy
is 97.79\%, 76.69\%, 85.87\%, 86.07\%, and 33.92\% for
\finalonly{}, \allprefix{}, \sampledprefix{}, \matchedexit{}, and
\progressive{}, respectively. At $r=192$, the corresponding means are
17.12\%, 72.72\%, 60.61\%, 53.71\%, and 17.19\%.
These readout-dependent differences and the seed variation in
Table~\ref{tab:control-h64} limit any inference from H32 retention to length
generalization. This follow-up evaluates H64 only; it does not establish a
new sampled length boundary over a range of longer horizons.

\subsection{Historical six-seed endpoint accuracy}
\label{app:endpoint}
This subsection preserves the original seeds 9--14 on stream 20260909.
Its results describe a different evaluation cohort from
Tables~\ref{tab:control-h32}--\ref{tab:control-h64}, even where model seeds
overlap.
\begin{table}[t]
 \centering
 \caption{\textbf{Paired endpoint accuracy beyond the training exit.}
 $A(32,r)$ in percent, averaged over six model seeds 9--14, 512 examples per
 seed, evaluation stream 20260909.
 The input length, sample stream, and metric are fixed across columns.}
 \label{tab:budget}
 \small
 \begin{tabular}{lrrrrrr}
  \toprule
  Inference loops $r$ & 32 & 33 & 36 & 40 & 48 & 64\\
  \midrule
  \finalonly & 100.00 & 74.90 & 52.41 & 39.78 & 33.01 & 30.50\\
  \allprefix & 100.00 & 100.00 & 100.00 & 100.00 & 100.00 & 97.85\\
  \bottomrule
 \end{tabular}
\end{table}
Table~\ref{tab:endpoint-seeds} gives the two over-budget slices that bracket
the main comparison. All twelve models have endpoint joint accuracy 1.000 at
the trained exit 32. The table makes clear that the six-seed mean at loop 64
does not imply identical behavior in every retained model.

\begin{table}[ht]
 \centering
 \caption{\textbf{Endpoint joint accuracy by model seed.} Entries are
 percentages on 512 length-32 examples per model, evaluation seed 20260909.
 The model is frozen while the inference budget changes.}
 \label{tab:endpoint-seeds}
 \small
 \begin{tabular}{lrrrrrr}
  \toprule
  Model seed & 9 & 10 & 11 & 12 & 13 & 14\\
  \midrule
  \finalonly, $r=33$ & 56.25 & 73.44 & 51.56 & 94.14 & 88.48 & 85.55\\
  \allprefix, $r=33$ & 100.00 & 100.00 & 100.00 & 100.00 & 100.00 & 100.00\\
  \finalonly, $r=64$ & 33.40 & 23.83 & 26.37 & 31.05 & 36.52 & 31.84\\
  \allprefix, $r=64$ & 100.00 & 100.00 & 87.30 & 99.80 & 100.00 & 100.00\\
  \bottomrule
 \end{tabular}
\end{table}

\begin{figure}[t]
 \centering
 \includegraphics[width=\linewidth]{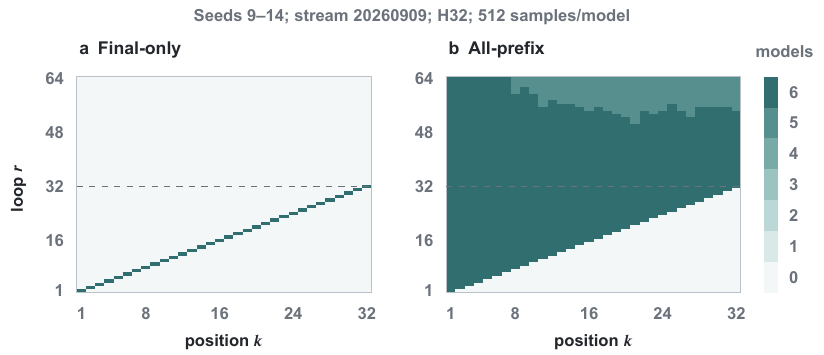}
 \caption{\textbf{Original readability grids for the serial baselines.}
 This historical view uses seeds 9--14 and evaluation stream 20260909;
 each cell counts models whose state and output marginal accuracies at that
 position and loop are both at least 99\%. This criterion differs from sample-level
 whole-trajectory exactness in Figure~\ref{fig:behavior}.
 It is retained separately from the common-stream training comparison.}
 \label{fig:readability-legacy}
\end{figure}

\subsection{Budget contracts and training support}
\label{app:budget}
A training contract assigns the number of repeated blocks to a sampled trace
length. All examples in a training batch share that sampled horizon and loop
count. Table~\ref{tab:contracts} specifies the relevant contracts. Stochastic
counts are drawn during training; their midpoint is used for a single-budget
evaluation when no count is explicitly requested. Budget sweeps instead
specify every evaluated loop count.

\begin{table}[ht]
 \centering
 \caption{\textbf{Training budget functions.} $U$ is discrete uniform.
 Slack-4 and Jitter-8 have the same conditional expected budget; the other
 rows do not form an equal-compute comparison.}
 \label{tab:contracts}
 \small
 \begin{tabular}{lll}
  \toprule
  Contract & Training budget & Interpretation at $H=32$\\
  \midrule
  Serial & $R=H$ & Trained exit 32\\
  Double & $R=2H$ & Trained exit 64\\
  Slack-4 & $R=H+4$ & Trained exit 36\\
  Slack-8 & $R=H+8$ & Trained exit 40\\
  Jitter-8 & $R=H+U\{0,\ldots,8\}$ & Trained support 32--40\\
  Random-2x & $R=H+U\{0,\ldots,H\}$ & Trained support 32--64\\
  \bottomrule
 \end{tabular}
\end{table}

\paragraph{Matching recurrent work.}
The Slack-4/Jitter-8 comparison fixes the architecture, final-only supervision,
optimizer, curriculum, data seed, batch size 256, and 132,000 training steps.
Each condition contains all six model seeds 9--14. Both have
$\mathbb{E}[R\mid H]=H+4$. Replaying the training functions with their recorded
random seeds gives the same sampled horizon sequence for both conditions.
Define $C_{\mathrm{batch}}=\sum_{t=1}^{132000}R_t$: this counts applications of
the shared block to a complete batch, accumulated over the entire training
run. It is not the loop count of one batch. The realized totals are
2,123,041 for Slack-4 and 2,121,730 for Jitter-8, a 0.06175\% reduction for
Jitter-8. The budget schedule is shared across model seeds. These counts
measure recurrent block work, not complete training FLOPs: they are not
weighted by actual padded sequence lengths and do not include embedding,
readout, or optimizer costs. For comparison, Serial and Double accumulate
1,595,041 and 3,190,082 batch block applications under this curriculum;
matching their optimizer steps does not match their recurrent work.

\paragraph{Endpoint retention and its tradeoffs.}
Table~\ref{tab:contract-results} evaluates the frozen Slack-4/Jitter-8 models
on the same length-32 stream, evaluation seed 20260909, with 512 examples per
model. Every trained seed is included, including those that did not master
the training length range. Jitter-8 has lower mean accuracy at the common
exit 36 but retains much higher accuracy with further updates. Loops 48, 64,
and 96 are outside both conditions' H32 training support. The per-seed
loop-64 slice in Table~\ref{tab:contract-seeds} shows that the retention
advantage occurs in all six pairs, rather than arising from selection of
successful Jitter-8 runs.

\begin{table}[ht]
 \centering
 \caption{\textbf{Endpoint retention at nearly matched recurrent work.}
 H32 endpoint joint accuracy in percent, averaged over all six paired seeds;
 512 examples per model, evaluation seed 20260909.}
 \label{tab:contract-results}
 \small
 \begin{tabular}{lrrrr}
  \toprule
  Inference loops $r$ & 36 & 48 & 64 & 96\\
  \midrule
  Slack-4 & 93.98 & 38.15 & 30.01 & 26.79\\
  Jitter-8 & 87.27 & 87.17 & 85.55 & 79.79\\
  \bottomrule
 \end{tabular}
\end{table}

\begin{table}[ht]
 \centering
 \caption{\textbf{The loop-64 retention comparison by seed.}
 H32 endpoint joint accuracy in percent, using the same evaluation stream
 and population size as Table~\ref{tab:contract-results}.}
 \label{tab:contract-seeds}
 \small
 \begin{tabular}{lrrrrrr}
  \toprule
  Model seed & 9 & 10 & 11 & 12 & 13 & 14\\
  \midrule
  Slack-4 & 32.23 & 31.25 & 33.79 & 34.96 & 15.04 & 32.81\\
  Jitter-8 & 95.51 & 100.00 & 100.00 & 57.23 & 60.55 & 100.00\\
  \bottomrule
 \end{tabular}
\end{table}

The wider valid-readout window does not imply uniformly better learning or
length generalization. On the separate training-log evaluation stream
(seed 20260902), five of six Slack-4 models and four of six Jitter-8 models
reach .99 endpoint joint accuracy at every evaluated horizon 1--32. On the
budget-sweep stream (seed 20260909), mean H64 endpoint joint accuracy at the
common readout $r=68$ is 74.32\% for Slack-4 and 64.36\% for Jitter-8. These
length-OOD scores and the ID success counts use distinct evaluation streams;
they are not pooled with the H32 retention measurements as one sample set.

The additional contracts provide different controls. The six audited Double
models have $\tau(k)=2k-1$ and observed holds of two to five loops, but this
comparison includes twice the recurrent work of Serial. Slack-8 does not
match Jitter-8's expected budget. Random-2x successes at H32 exits 32--64 are
inside its training support and do not establish generalization to unseen
loop counts; a deterministic mean-budget control for Random-2x is absent.
These observations delimit what the budget experiments can attribute to the
training exit distribution.

\subsection{Whole-trajectory exactness and complementary contracts}
\label{app:trajectory}
Endpoint joint accuracy asks whether the final readout is correct. A stricter
sample-level quantity asks whether every one of the $H=32$ action positions has
both its state and output correct. We call a sample satisfying this condition
\emph{trajectory exact}. This is not the same as the per-position marginal
threshold used by the readability grids in Figure~\ref{fig:readability-legacy}: one
incorrect position makes a sample trajectory-inexact.

\begin{table}[ht]
 \centering
 \caption{\textbf{Whole-trajectory exactness after the H32 exit.}
 Percent of 512 length-32 examples, averaged over six model seeds 9--14;
 evaluation stream 20260909. Every action position must have correct state
 and output. The endpoint means at $r=64$ are shown in
 Tables~\ref{tab:budget} and~\ref{tab:contract-results}; the two metrics measure different
 failure modes.}
 \label{tab:trajectory-exact}
 \small
 \begin{tabular}{lrr}
  \toprule
  Condition & $r=64$ & $r=96$\\
  \midrule
  \finalonly & 0.07 & 0.07\\
  \allprefix & 86.39 & 72.46\\
  Slack-4 & 0.00 & 0.00\\
  Jitter-8 & 54.46 & 47.40\\
  \bottomrule
 \end{tabular}
\end{table}

The same separation already appears at the trained exit: both Serial
conditions have 100\% endpoint accuracy at $r=R=32$, whereas trajectory exactness
is 0.10\% for \finalonly{} and 100\% for \allprefix{}. Thus the all-prefix result
is not only a more reliable final readout; it preserves a substantially larger
fraction of complete recurrent trajectories. For the matched-budget contracts,
Jitter-8 similarly retains complete trajectories after the nominal support,
while Slack-4 does not. These figures are descriptive paired-cohort means and
do not establish that randomization is uniformly better inside the training
support.

For context, two additional contracts delimit the interpretation of the
retention comparison. Slack-8 is a deterministic extra-slack dose control: its
H32 endpoint means are 100.00\% at its trained exit $R=40$, then 92.74\%,
57.94\%, and 44.21\% at $r=48,64,96$, respectively (six seeds, same stream
and model recipe). It does not match Jitter-8's expected budget. The audited
Double cohort instead trains at $R=2H$: on H32 it reaches 100\% at its trained
exit $R=64$, but only 2.93\% when forced to read at $r=32$ (ten seeds). This
positive control shows that long loops can be learned when they are the
training contract; it is not a matched-compute estimate of the retention
effect.

\subsection{Length extrapolation}
\label{app:length}
The sampled length boundary $B_{\mathrm{final}}$ is the largest evaluated
horizon whose last-position joint accuracy is at least .99, under the model's
own serial contract $R=H$. It is a maximum over sampled horizons, not a proof
of correctness at every intermediate or greater length. Table~\ref{tab:length}
reports the six paired values. The retention objective improves this metric
in three seeds and reduces it in three, despite improving H32 answer
retention. This does not establish that retention and length generalization
are unrelated; it establishes that the observed retention gain does not
provide a seed-wise guarantee of an improved sampled boundary.

\begin{table}[ht]
 \centering
 \caption{\textbf{The paired sampled length boundary has mixed changes.}
 $B_{\mathrm{final}}$ is defined by final-position joint accuracy, not by
 all-position trajectory exactness.}
 \label{tab:length}
 \small
\begin{tabular}{lrrrrrr}
\toprule
Seed & 9 & 10 & 11 & 12 & 13 & 14 \\
\midrule
\finalonly & 88 & 56 & 72 & 56 & 48 & 56 \\
\allprefix & 72 & 88 & 48 & 48 & 64 & 80 \\
\bottomrule
\end{tabular}

\end{table}

\subsection{Delayed swaps with their clean controls}
\label{app:delayed}
The six paired \finalonly\ and \allprefix\ models of the original cohort
receive interventions at position 16 of H32 inputs. The 256 different-state
pairs are shared across models; 242 have distinguishable counterfactual
targets. This population is fixed throughout, without selection on neural
success.

At position 16 and loop 16, clean suffix exactness is 1.000 along the
frontier in both conditions; source suffix exactness is 1.000 versus .939
(\finalonly\ versus \allprefix). At exit 32, clean exactness is .007 versus
1.000 and source exactness is .002 versus .722. Thus \allprefix\ supervision
improves the common-exit readout of an early intervention, without improving
frontier transport. Delaying both source collection and insertion by $a$
loops, and reading position $k$ at $k+a$, holds the number of
post-intervention updates fixed. Under \allprefix, clean suffix exactness
stays at 1.000, but source exactness falls from .939 to .177, .102, and .000
at ages 1, 2, and 4. Even inserting the original loop-16 source at loop 20
fails: joint agreement with the position-17 counterfactual at loop 21 is 0\%,
compared with 94.63\% when inserted at loop 16 and read at loop 17, while the
late clean base run remains correct (Appendix~\ref{app:phase}).

\begin{figure}[t]
 \centering
 \includegraphics[width=\linewidth]{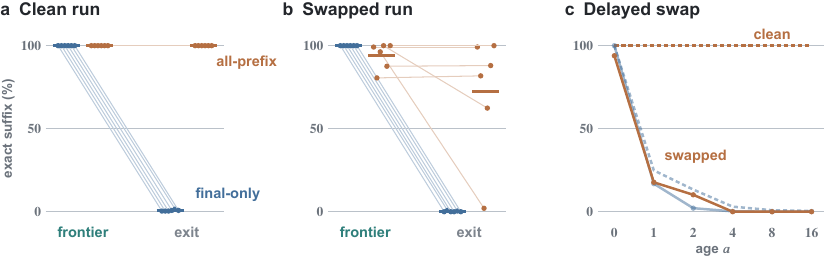}
 \caption{\textbf{Retention training changes the two readouts of an early
 swap differently, but does not protect a delayed swap.} Swap at position 16;
 six paired models per condition. Exact suffix requires every state and
 output at positions 17--32 to match the target. (a) Unswapped runs scored
 against their own trajectory, all 256 pairs. (b) Swapped runs scored
 against the source continuation, 242 eligible pairs. Frontier reads position
 $k$ at loop $k$; exit reads every position at loop 32. Lines join the two
 readouts of one model; bars are means. (c) The source vector is taken and
 inserted $a$ loops after the frontier, and position $k$ is read at loop
 $k+a$ (solid), next to the matching unswapped run (dashed); means over six
 models. \finalonly\ is faded because its unswapped runs are no longer
 readable at $k+a$. Clean and source scores use different targets and
 denominators and are not subtracted.}
 \label{fig:supervision}
\end{figure}
Table~\ref{tab:delayed} places clean and source suffix exactness next to each
other for the main delayed ages. Clean results score the unmodified
base run's own suffix over all 256 pairs; swap results score the source
counterfactual on the fixed 242 eligible pairs. The two columns should not
be subtracted as an effect estimate for a common target and denominator.
Exit-48 values differ from exit-32 values by less than .036 across the reported
condition/age entries. Unlike the shifted-frontier and exit-32 schedules,
\allprefix\ clean suffix exactness at exit 48 is slightly below 1.000
(six-seed mean .9961; seed 11, .9766). This additional fixed exit does not remove the
source-phase and base-phase changes inherent in delayed insertion.

\begin{table}[ht]
 \centering
 \caption{\textbf{Ordinary retention is essential to interpreting a delayed
 swap.} Exactness of positions 17--32, averaged over six paired models. The
 shifted-frontier schedule reads position $k$ at $k+a$; the exit schedule reads
 every position at loop 32.}
 \label{tab:delayed}
 \small
\begin{tabular}{lrrrrr}
\toprule
 & & \multicolumn{2}{c}{\finalonly} & \multicolumn{2}{c}{\allprefix} \\
\cmidrule(lr){3-4}\cmidrule(lr){5-6}
Readout & $a$ & No swap & Swap & No swap & Swap \\
\midrule
Frontier & 0 & 1.000 & 1.000 & 1.000 & .939 \\
 & 1 & .249 & .169 & 1.000 & .177 \\
 & 2 & .133 & .021 & 1.000 & .102 \\
 & 4 & .031 & .002 & 1.000 & .000 \\
\addlinespace
Exit & 0 & .007 & .002 & 1.000 & .722 \\
 & 1 & .007 & .001 & 1.000 & .074 \\
 & 2 & .007 & .001 & 1.000 & .105 \\
 & 4 & .007 & .001 & 1.000 & .000 \\
\bottomrule
\end{tabular}

\end{table}

\subsection{Crossed times and replacement extent}
\label{app:phase}
Table~\ref{tab:phase} records the primary time-crossing slice from
Appendix~\ref{app:delayed}. Each entry scores one target position and should not
be compared as if it were complete-suffix exactness. At the early readout,
clean factual correctness is 1.000 under both conditions; at the late readout
it is .385 for \finalonly\ and 1.000 for \allprefix. The late \finalonly\
base run's unmodified agreement with the source target is .167. These
baselines prevent a raw late score from being mistaken for transferred
information.

\begin{table}[ht]
 \centering
 \caption{\textbf{Source collection and base insertion at crossed times.}
 All vectors are at position 16. Position 17 is read one update after insertion,
 using the same 242 eligible pairs and six model seeds. Entries are joint
 state--output agreement with the source target.}
 \label{tab:phase}
 \small
 \begin{tabular}{llrr}
  \toprule
  Source loop & Base loop & \finalonly & \allprefix\\
  \midrule
  16 & 16 & 1.000 & .946\\
  20 & 16 & .196 & .260\\
  16 & 20 & .267 & .000\\
  20 & 20 & .169 & .000\\
  \bottomrule
 \end{tabular}
\end{table}

The extent comparison in Table~\ref{tab:extent} asks where an effective
replacement can be made at the common loop 20. Its scored target, position 21,
is never directly replaced. The downstream-only block includes position 20,
which is the current frontier. Therefore its success supplies a localized
positive reference, but neither it nor the larger block demonstrates that
position 16's old vector alone is sufficient. The old site's extra distance to
the target also prevents an equal-distance inference of state erasure.

\begin{table}[ht]
 \centering
 \caption{\textbf{Replacement extent at loop 20.} Position 21 is read at loop
 21, on 121 eligible pairs. Both clean base runs are correct on every pair.
 Unless specified, all source vectors are collected at loop 20.}
 \label{tab:extent}
 \small
\begin{tabular}{lrr}
\toprule
Recipient vectors replaced at loop 20 & \finalonly & \allprefix \\
\midrule
Cycle 16, donor vector from loop 16 & .163 & .161 \\
Cycle 16 & .000 & .000 \\
Cycle 20 (current frontier) & 1.000 & .935 \\
Cycles 16--20 & 1.000 & .934 \\
Cycles 17--20 & 1.000 & .934 \\
\bottomrule
\end{tabular}

\end{table}

\subsection{Provenance of the principal quantities}
\label{app:provenance}
The supplementary provenance index records each paired budget curve and
original swap summary with its model seed, evaluation seed, checkpoint step,
exact source path, SHA256 hash, and per-seed output. Twelve
paired budget-sweep files and ten original swap files matched byte-for-byte
between the local and remote evidence stores. The paired retention tables
derive from the completed P1 result set, and the phase/extent tables from
the validated compatibility follow-up. These later experiments use distinct
pair populations and scoring slices from the original position-8 audit. They
are not pooled into its numerator or denominator.

Low-dimensional alignment and bounded automaton extraction were explored in
the wider project, but do not enter the present paper's claims. The reported
conclusions depend on the fixed-checkpoint behavioral evaluations, full-vector
interventions, and paired training comparison described above.

\section{Local downstream recovery and source-phase controls}
\label{app:receiver}

The phase-crossing experiment changes the base run's entire execution
phase. Here we keep the late background fixed and test whether changing one
local downstream vector can restore the response to a fresh source. The
protocol was fixed before inspecting this follow-up's outputs; it did not
retrospectively preregister the preceding experiments.

\subsection{Population, interventions, and scoring}

We evaluated all twelve frozen step-132,000 checkpoints: six \finalonly\ and
six \allprefix\ models, with paired seeds 9--14. The experiment used $H=32$
and the position-16 pair population, with 256 different-state pairs and 256
same-state/different-history controls per model. Source and base runs shared
the machine, presentation, and future actions. The primary population was
the fixed set of 242 different-state pairs whose symbolic state--output
targets at position 17 differ. This selection used no neural predictions and
was identical across arms and models. Same-state pairs had no distinguishable
counterfactual target and were scored separately against their factual target.

Let $h_{k,S}^r$ and $h_{k,B}^r$ denote source and base vectors from
\emph{unmodified} runs. After loop 20, a single post-block,
pre-final-normalization hook performed a $2\times2$ intervention:
\begin{equation}
 \begin{aligned}
  i=1 &: \quad h_{16}^{20}\leftarrow h_{16,S}^{16},\\
  j=1 &: \quad h_{17}^{20}\leftarrow h_{17,B}^{16}.
 \end{aligned}
 \label{eq:receiver-reset}
\end{equation}
An index of zero left the corresponding late base vector unchanged.
Thus $(i,j)=(0,0)$ was clean, $(1,0)$ inserted a fresh source, $(0,1)$ reset
only the downstream position, and $(1,1)$ combined both changes. Both replacements in
the combined arm were simultaneous. All other positions, including the
table and remaining action positions, retained their loop-20 vectors.
The reset therefore changed the scored position's \emph{pre-update} hidden
state, using the base run's own earlier vector; it did not insert a source
position-17 vector or a symbolic answer. We ran one further shared-block update
and read position 17 at loop 21.

The primary outcome was joint counterfactual accuracy: both predicted state
and output at that position had to match the symbolic source continuation.
It was neither factual accuracy nor suffix exactness. A fresh positive
reference inserted $h_{16,S}^{16}$ at position 16 and loop 16 of the base run and read position
17 at loop 17. Secondary arms substituted the aged source $h_{16,S}^{20}$ at
loop 20, with and without the same downstream reset.

\begin{figure}[t]
 \centering
 \includegraphics[width=\linewidth]{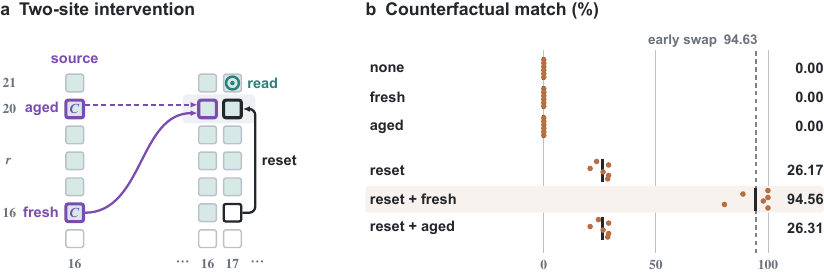}
 \caption{\textbf{Resetting the downstream position restores the use of a fresh source,
 but not of an equally readable aged source.} (a) The base run is held at
 loop 20 (gray band). Position 16 may receive the source's position-16 vector
 from loop 16 (fresh, solid) or loop 20 (aged, dashed), and position 17 may be
 reset to the base run's own loop-16 vector; other positions keep their
 loop-20 vectors. After one update, position 17 is scored at loop 21
 ($\odot$). Shaded cells are readable under \allprefix; both source vectors
 decode the correct state and output on every pair. (b) Rows name the
 replacements at loop 20 (none: unmodified run). Percentage of the 242
 eligible pairs whose position-17 state and output both match the
 counterfactual target, not factual accuracy or suffix exactness; dots are six
 \allprefix\ models, bars their means, and the dashed line the early swap.
 \finalonly\ results are in Table~\ref{tab:receiver-results} and controls in
 Table~\ref{tab:receiver-controls}.}
 \label{fig:receiver}
\end{figure}

\subsection{Joint restoration recovers the fresh response}

\begin{table}[t]
\centering
\small
\caption{\textbf{Single-step counterfactual accuracy after downstream recovery.}
All values are percentages, reported as mean $\pm$ sample standard deviation
over six model seeds. Each model uses the same 242 eligible pairs. Late arms
intervene at loop 20 and read position 17 at loop 21; the fresh reference
intervenes at loop 16 and reads at loop 17.}
\label{tab:receiver-results}
\begin{tabular}{lrr}
\toprule
Intervention & \finalonly & \allprefix\\
\midrule
Fresh source reference & $100.00\pm0.00$ & $94.63\pm8.14$\\
\midrule
Late clean ($A_{00}$) & $16.67\pm3.64$ & $0.00\pm0.00$\\
Late fresh source only ($A_{10}$) & $26.72\pm26.17$ & $0.00\pm0.00$\\
Late downstream reset only ($A_{01}$) & $19.63\pm2.97$ & $26.17\pm3.40$\\
Late fresh source + reset ($A_{11}$) & $100.00\pm0.00$ & $94.56\pm8.10$\\
\midrule
Late aged source only & $16.87\pm3.77$ & $0.00\pm0.00$\\
Late aged source + reset & $19.70\pm3.02$ & $26.31\pm3.29$\\
\bottomrule
\end{tabular}
\end{table}

Table~\ref{tab:receiver-results} and Figure~\ref{fig:receiver} show that the combined intervention closely
recovers the fresh reference under both supervision conditions. The
\allprefix\ late clean arm has zero counterfactual accuracy because it
correctly retains the distinct factual answer; its factual joint accuracy
on these pairs is 100\%. Neither a fresh source alone nor a downstream reset
alone explains the combined result. For each model, we computed the paired
interaction on the same eligible population,
\begin{equation}
 I=A_{11}-A_{10}-A_{01}+A_{00}.
 \label{eq:receiver-interaction}
\end{equation}
The mean $I$ was $+70.32\pm22.39$ percentage points under \finalonly\ and
$+68.39\pm10.54$ under \allprefix, and was positive in all six models of
each condition. These are descriptive within-model contrasts, with model
seeds as the replication unit, rather than significance tests that pool
reused task pairs as independent replications. Table~\ref{tab:receiver-seeds}
also exposes the substantial seed variation in \finalonly\ source-only
performance.

\begin{table}[t]
\centering
\small
\caption{\textbf{Per-seed primary outcomes.} Entries are successful joint
counterfactual predictions out of 242 eligible pairs for every cell.
``Source'' means the fresh loop-16 source inserted at loop 20.}
\label{tab:receiver-seeds}
\begin{tabular}{llrrrrrr}
\toprule
Condition & Late intervention & 9 & 10 & 11 & 12 & 13 & 14\\
\midrule
\finalonly & Clean & 53 & 37 & 35 & 34 & 33 & 50\\
 & Source only & 193 & 35 & 45 & 33 & 31 & 51\\
 & Reset only & 41 & 57 & 52 & 51 & 38 & 46\\
 & Source + reset & 242 & 242 & 242 & 242 & 242 & 242\\
\midrule
\allprefix & Clean & 0 & 0 & 0 & 0 & 0 & 0\\
 & Source only & 0 & 0 & 0 & 0 & 0 & 0\\
 & Reset only & 69 & 70 & 64 & 50 & 70 & 57\\
 & Source + reset & 242 & 195 & 237 & 242 & 215 & 242\\
\bottomrule
\end{tabular}
\end{table}

Recovery extends beyond matching aggregate rates. Comparing the predicted
state--output pair under the fresh reference and late combined intervention
gives identical predictions for all 3,072 \finalonly\ model--pair evaluations
($6$ seeds, $2$ pair kinds, $256$ pairs), and all but one of the 3,072
\allprefix\ evaluations. The exception is a different-state pair in seed 11
that succeeds under the fresh reference and fails after late restoration.
This establishes agreement of discrete single-step behavior; it does not
establish equality of the hidden states.

\subsection{A readable aged source does not provide the same recovery}

Source readability was measured at the actual source position used in the
intervention, with the original state and output heads. Fresh sources were
jointly correct on all different-state pairs in both conditions. Under
\allprefix, aged sources were also 100\% correct for both heads in every
seed, over both the full 256-pair population and the 242 eligible pairs.
Nevertheless, aged source plus reset reached only 26.31\%, almost identical
to reset alone at 26.17\%. Its net gain was two successful model--pair
evaluations across six sets of 242; the corresponding aged-source
interaction was $+0.14\pm0.34$ percentage points. The 26.31\% level is
therefore largely the reset baseline, not evidence of effective aged-source
transport. Under \finalonly, the aged-source interaction was
$-0.14\pm0.34$ points, but aged source joint readability was only 39.88\%
on the eligible pairs. The cleaner evidence for readable yet
non-interchangeable sources comes from \allprefix.

\subsection{Controls, reproducibility, and interpretation}

\begin{table}[t]
\centering
\small
\caption{\textbf{Same-state/different-history controls.} Factual joint
accuracy in percent, mean $\pm$ sample standard deviation over six seeds,
using 256 pairs per model. Source and factual symbolic continuations coincide,
so distinguishable counterfactual accuracy is undefined for this population.}
\label{tab:receiver-controls}
\begin{tabular}{lrr}
\toprule
Intervention & \finalonly & \allprefix\\
\midrule
Fresh clean & $100.00\pm0.00$ & $100.00\pm0.00$\\
Fresh source reference & $100.00\pm0.00$ & $95.18\pm6.49$\\
Late clean & $47.27\pm15.78$ & $100.00\pm0.00$\\
Late fresh source only & $57.23\pm18.05$ & $100.00\pm0.00$\\
Late downstream reset only & $33.46\pm8.66$ & $25.13\pm1.18$\\
Late fresh source + reset & $100.00\pm0.00$ & $95.18\pm6.49$\\
Late aged source only & $47.20\pm15.79$ & $100.00\pm0.00$\\
Late aged source + reset & $33.53\pm8.66$ & $25.20\pm1.18$\\
\bottomrule
\end{tabular}
\end{table}

Same-state controls show that the downstream reset is a strong intervention:
\allprefix\ factual accuracy falls from 100\% to 25.13\% with reset alone
(Table~\ref{tab:receiver-controls}). Fresh source plus reset restores 95.18\%,
matching the fresh reference and its roughly 4.8\% substitution damage.
For different-state eligible pairs, reset alone yields 26.17\% source-correct,
20.94\% factual-correct, and 52.89\% matching neither target. Thus resetting
does not itself provide a general accuracy repair, and complete-vector
substitution cannot be interpreted as exchanging only a pure symbolic state.

True same-phase self-replacements were separately tested at position 16
with the fresh source, position 16 with the late source, late position 17, and both late positions. All preserved
the corresponding clean predictions; saved pre-intervention invariants
also passed. The evaluation reused stream 2026091001 in batches of 16,
with batch streams formed by adding
$10000019\cdot\text{kind index}+1000003\cdot\text{batch index}$.
The separate 16-pair-per-kind execution pilot was excluded from all reported
statistics. The experiment records contain the frozen protocol, checkpoint
identities, pair identities, raw predicted and symbolic labels, and aggregate
results. An independent scoring audit recomputed source rates and denominators
from the recorded labels, without importing the aggregation implementation,
and reproduced the reported results.

The positive result localizes a causal contribution of the downstream state:
while the remaining network stays at its late phase, restoring one downstream
vector together with a fresh source nearly reproduces the earlier
single-step response. The aged-source comparison shows that downstream
restoration alone does not make every readable source interchangeable.
These conclusions are restricted to $H=32$, insertion at position 16, reset of position 17, insertion
at loop 20, and one subsequent update. Full-vector replacements can restore
readiness, representation format, or other aspects of local compatibility;
they do not identify a particular attention head, gate, or minimal state
code. The experiment establishes neither information erasure nor necessity
in unmodified execution, and does not demonstrate recovery of a whole
suffix, length extrapolation, or the cause of natural endpoint deterioration
from loop 32 to 33. That failure is localized by a separate experiment
(Appendix~\ref{app:natural}).

\section{Frontier freezing}
\label{app:freeze}

\subsection{Original cohort and protocol}
The original freezing cohort uses the frozen checkpoints of Section~\ref{sec:temporal} on a new
evaluation stream (20260923), 512 inputs per model and horizon, with the
same inputs for every arm of a given model. The schedule is fixed before
scoring: models trained at $R=H$ use $s(k)=k$. \emph{Frontier freezing} captures
position $k$'s post-block hidden vector at loop $s(k)$ on the arm's own
ongoing trajectory and restores it after every later block. \emph{One loop
later} captures at $s(k)+1$. \emph{Positions 1--8} freezes only those
positions, so every later position must still compute. \emph{Readout cache}
runs the clean forward pass and delivers each position's prediction saved at
$s(k)$, without modifying any hidden state. No vector comes from a
separate run, no label enters an intervention, and table and presentation
tokens are never frozen. Frozen positions remain available to attention.
All layers of a block still execute; freezing acts at recurrent block
boundaries, not on individual layers or key--value caches. The protocol was
written before the results were inspected but is exploratory, not an
independent confirmation.

\begin{table}[h]
 \centering
 \footnotesize
 \setlength{\tabcolsep}{4pt}
 \caption{Endpoint joint accuracy (\%) at loop 32 for $H=32$, 512 inputs per
 model. The larger models use $d=512$ and four layers per block and a batch
 size of 32, so they are not paired input by input with the other rows.}
 \label{tab:freeze}
 \begin{tabular}{llrrrrr}
 \toprule
 Condition & Seed & Clean & Readout cache & Frontier freeze & Positions 1--8 & One loop later \\
 \midrule
 \finalonly & 9 & 100.0 & 100.0 & 11.5 & 11.9 & 100.0 \\
 \finalonly & 10 & 100.0 & 100.0 & 6.2 & 8.0 & 100.0 \\
 \finalonly & 11 & 100.0 & 100.0 & 1.6 & 5.1 & 100.0 \\
 \finalonly & 12 & 100.0 & 100.0 & 1.6 & 1.8 & 100.0 \\
 \finalonly & 13 & 100.0 & 100.0 & 2.7 & 5.5 & 77.1 \\
 \finalonly & 14 & 100.0 & 100.0 & 1.6 & 1.6 & 100.0 \\
 \midrule
 \allprefix & 9 & 100.0 & 100.0 & 3.7 & 3.7 & 100.0 \\
 \allprefix & 10 & 100.0 & 100.0 & 0.8 & 0.6 & 94.1 \\
 \allprefix & 11 & 100.0 & 100.0 & 16.0 & 29.9 & 75.0 \\
 \midrule
 \finalonly, $d{=}512$ & 9 & 100.0 & 100.0 & 8.2 & 7.8 & 100.0 \\
 \finalonly, $d{=}512$ & 10 & 100.0 & 100.0 & 3.1 & 3.3 & 99.0 \\
 \finalonly, $d{=}512$ & 11 & 100.0 & 100.0 & 12.9 & 18.0 & 74.4 \\
 \bottomrule
 \end{tabular}
\end{table}

\subsection{Additional observations and limits}
Table~\ref{tab:freeze} reports every evaluated model. Under frontier
freezing, positions 1 and 2 remain correct and accuracy at each position's
scheduled capture loop falls from position 3 onward in all three groups. Capturing one
loop later is not uniformly harmless. It lowers the endpoint to 74--77\% in
three of the twelve models at $H=32$. On the separate $H=64$ length
evaluation at readout $r=H=64$, it lowers the endpoint
relative to clean in several \finalonly\ models (seed 12: 44.1\% versus
99.2\%; seed 14: 46.3\% versus 95.7\%). One-loop-later freezing is therefore
a timing contrast, not a safe schedule. \double\ models, whose observed clean frontier is
$\tau(k)=2k-1$, collapse under both schedules; their clean endpoint at $r=H$
is already low because they need about $2H$ loops, so they are not
interpreted. The readout cache preserves the captured predictions by
construction and is correct on every original H32 example. It is the
baseline any freezing schedule would have to beat; we propose no such
schedule.

\begin{figure}[t]
 \centering
 \includegraphics[width=\linewidth]{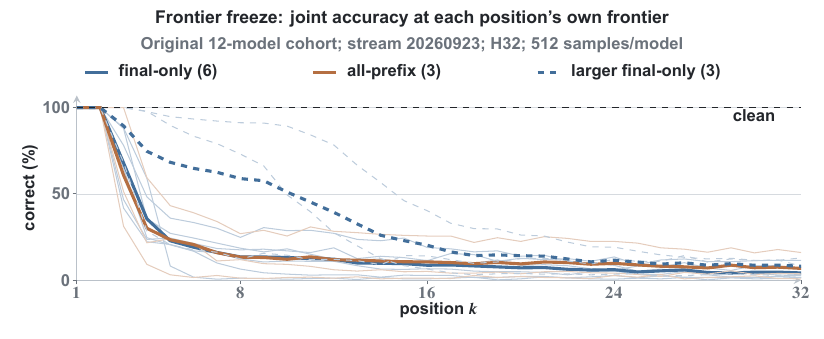}
 \caption{\textbf{Where continuing computation fails in the original
 freezing cohort.} Joint state--output accuracy at each position's
 scheduled readout on H32, stream 20260923. Thick curves are means within
 the three original model groups and thin curves show individual seeds.
 Positions that remain readable early in the sequence do not prevent
 failure at later positions under immediate freezing.}
 \label{fig:freeze-positions}
\end{figure}

\subsection{Freezing the new training controls}
\label{app:freeze-controls}
This follow-up evaluates all three \sampledprefix{} and all three
\matchedexit{} checkpoints from Appendix~\ref{app:training-controls},
at step 132,000, without retraining. It was specified after inspecting
their clean retention results and before inspecting their freeze results;
it is exploratory. The fresh stream is 20260924, distinct from the
20260923 training-control evaluation. Each model uses 512 H32 examples in
four batches of 128, evaluated for 64 loops. Saved input hashes match
across all six models. Every model and all 512 examples are retained,
without filtering on clean correctness. The two runtime pilots use a
different stream and are excluded. \progressive{} is outside this bounded
follow-up: two models had not learned H32 and the successful model had a
shifted clean frontier, requiring a separately specified schedule analysis.

\paragraph{Fixed schedules and eight arms.}
The reference schedule $s(k)=k$ is fixed before evaluation and is not
estimated from the new labels. The evaluator and its interventions are
unchanged. Besides clean computation and readout caching, all-position
freezing captures at $s_d(k)=k+d$ for each $d\in\{0,1,2,4\}$; two prefix arms
capture only positions 1--8 or 1--16 at $k$. Each capture uses the arm's
own current vector after a whole block and restoration first affects the
next loop. All remaining positions and presentation tokens stay live;
attention can still use the held vectors. These interventions do not save
FLOPs, use hidden states from another run, or establish layerwise or
key--value cache freezing.

\begin{table}[t]
 \centering
 \small
 \setlength{\tabcolsep}{4pt}
 \caption{\textbf{Clean diagnostics for every new freeze model.} H32,
 stream 20260924, 512 inputs per model; percentages. $A_{32}$ and $T_{32}$
 denote endpoint and whole-trajectory accuracy at loop 32. ``Min.
 diagonal'' is the minimum over positions of clean joint accuracy at
 $r=k$. Cache $T_{32}$ requires every saved prediction to be correct.}
 \label{tab:control-freeze-diagnostics}
\begin{tabular}{lrrrrr}
\toprule
Condition & Seed & Clean $A_{32}$ & Clean $T_{32}$ & Min. diagonal & Cache $T_{32}$ \\
\midrule
\sampledprefix & 9 & 100.00 & 100.00 & 100.00 & 100.00 \\
\sampledprefix & 10 & 100.00 & 100.00 & 100.00 & 100.00 \\
\sampledprefix & 11 & 100.00 & 100.00 & 100.00 & 100.00 \\
\matchedexit & 9 & 99.80 & 99.02 & 96.48 & 94.92 \\
\matchedexit & 10 & 100.00 & 32.03 & 99.80 & 99.80 \\
\matchedexit & 11 & 100.00 & 73.83 & 100.00 & 100.00 \\
\bottomrule
\end{tabular}

\end{table}

\paragraph{Clean eligibility.}
All three \sampledprefix{} models have exact clean and cached H32
trajectories on this stream, including every diagonal readout at $r=k$.
For \matchedexit{}, seed 11 also has exact diagonal reads; seed 10 has
99.80\% minimum diagonal accuracy and 99.80\% cached trajectory exactness.
Seed 9 has 96.48\% minimum diagonal accuracy, 94.92\% cached trajectory
exactness, and 99.80\% clean endpoint accuracy. Thus the reference schedule
$s(k)=k$ does not give perfectly correct clean readouts across this entire
cohort; it is prescribed independently of the measured threshold-crossing
loop $\tau(k)$.
The observed clean errors are reported as such;
they are not evidence of a correct state failing under freezing.
Even when the clean diagonal is exact, later vectors on an online-freeze
trajectory may already be incorrect because earlier positions have been
held. Clean readability at the reference schedule therefore does not imply
that every vector actually captured under intervention is correct.

\begin{table}[t]
 \centering
 \small
 \setlength{\tabcolsep}{4pt}
 \caption{\textbf{Endpoint accuracy at the scheduled capture readout.}
 Three-seed means in percent on all 512 H32 examples per model, stream
 20260924. The $+d$ columns read the endpoint at $r=32+d$; clean, cache,
 P8, and P16 use $r=32$. P8/P16 freeze only the indicated prefix.
 Clean endpoint accuracy at $r=33,34,36$ is 100\% for every model;
 the clean $r=32$ exception is reported in
 Table~\ref{tab:control-freeze-diagnostics}.}
 \label{tab:control-freeze-capture}
\begin{tabular}{lrrrrrrrr}
\toprule
Condition & Clean & Cache & $+0$ & $+1$ & $+2$ & $+4$ & P8 & P16 \\
\midrule
\sampledprefix & 100.00 & 100.00 & 4.10 & 80.21 & 99.80 & 100.00 & 7.49 & 4.36 \\
\matchedexit & 99.93 & 99.93 & 6.25 & 20.18 & 87.24 & 99.80 & 7.42 & 6.25 \\
\bottomrule
\end{tabular}

\end{table}

\paragraph{Capture-aligned and common readouts.}
For a delayed arm, the endpoint has not yet been captured at $r=32$.
We therefore use $r=32+d$ for the endpoint comparison in the main figure
and Table~\ref{tab:control-freeze-capture}, with clean evaluated at the
same loop. Common-loop scores at $r=32,33,64$ are retained in full below.
For example, \sampledprefix{} seed 10 scores 100\% at $r=32$ under
delay 1 but only 82.81\% when its endpoint is captured at $r=33$; clean
is 100\% at both loops. Scoring only $r=32$ would overstate the recovery.
Scheduled whole-trajectory exactness reads each position at its own
$k+d$ on the intervened trajectory. It differs from scoring every
position simultaneously at a common loop. For prefix arms, the scheduled
readout remains $r=k$, including positions that are never frozen.

\paragraph{Results and limits.}
Immediate freezing reduces mean endpoint accuracy to 4.10\% for
\sampledprefix{} and 6.25\% for \matchedexit{}, with zero
whole-trajectory exactness in every model. Thus the failure to hold
position states extends across these training objectives, including three
models with fully correct clean scheduled reads. Delaying capture helps,
but the required delay depends on the model and training condition.
At delay 2, \sampledprefix{} reaches 99.80\% mean endpoint accuracy and
99.54\% scheduled trajectory exactness. \matchedexit{} reaches only
87.24\% endpoint accuracy at delay 2, and needs delay 4 to reach 99.80\%
endpoint accuracy and 99.74\% scheduled trajectory exactness. All fixed
delays and all seeds are reported; one loop is not a generally safe
schedule. Prefix-only freezing also damages the never-frozen endpoint
and suffix, so the collapse is not limited to scoring held positions.
The two objectives differ in losses and training work, precluding a unique
attribution of their delay difference to one training factor. These
results concern holding a state on its own evolving trajectory; they do
not demonstrate arbitrary vector interchangeability, convergence, or a
particular microscopic circuit.

\begin{table}[p]
 \centering
 \small
 \setlength{\tabcolsep}{4pt}
 \caption{\textbf{Every \sampledprefix{} endpoint score under freezing.}
 Percent of 512 H32 examples, stream 20260924. Numerical readouts are
 common loops. ``Sched.'' reads the endpoint at $32+d$ for $+d$ and at
 32 for clean/cache/P8/P16. Cache leaves the hidden computation live;
 P8/P16 freeze only positions 1--8/1--16. All eight arms and all seeds
 are included.}
 \label{tab:control-freeze-sampled-endpoint}
\begin{tabular}{lrrrrrrrrr}
\toprule
Seed & Readout & Clean & Cache & $+0$ & $+1$ & $+2$ & $+4$ & P8 & P16 \\
\midrule
9 & $32$ & 100.00 & 100.00 & 3.91 & 100.00 & 100.00 & 100.00 & 3.91 & 3.71 \\
9 & $33$ & 100.00 & 100.00 & 3.91 & 100.00 & 100.00 & 100.00 & 3.91 & 3.91 \\
9 & $64$ & 100.00 & 100.00 & 3.91 & 100.00 & 100.00 & 100.00 & 4.10 & 4.10 \\
9 & Sched. & 100.00 & 100.00 & 3.91 & 100.00 & 100.00 & 100.00 & 3.91 & 3.71 \\
\midrule
10 & $32$ & 100.00 & 100.00 & 2.54 & 100.00 & 100.00 & 100.00 & 5.66 & 2.54 \\
10 & $33$ & 100.00 & 100.00 & 2.54 & 82.81 & 100.00 & 100.00 & 6.45 & 2.54 \\
10 & $64$ & 99.61 & 100.00 & 2.54 & 82.81 & 100.00 & 100.00 & 11.91 & 4.88 \\
10 & Sched. & 100.00 & 100.00 & 2.54 & 82.81 & 100.00 & 100.00 & 5.66 & 2.54 \\
\midrule
11 & $32$ & 100.00 & 100.00 & 5.86 & 98.83 & 100.00 & 100.00 & 12.89 & 6.84 \\
11 & $33$ & 100.00 & 100.00 & 5.86 & 57.81 & 99.61 & 100.00 & 12.50 & 6.84 \\
11 & $64$ & 99.41 & 100.00 & 5.86 & 57.81 & 99.41 & 100.00 & 16.41 & 7.23 \\
11 & Sched. & 100.00 & 100.00 & 5.86 & 57.81 & 99.41 & 100.00 & 12.89 & 6.84 \\
\bottomrule
\end{tabular}

\end{table}

\begin{table}[p]
 \centering
 \small
 \setlength{\tabcolsep}{4pt}
 \caption{\textbf{Every \matchedexit{} endpoint score under freezing.}
 Same complete reporting and readout definitions as
 Table~\ref{tab:control-freeze-sampled-endpoint}; 512 H32 examples per
 model, stream 20260924.}
 \label{tab:control-freeze-matched-endpoint}
\begin{tabular}{lrrrrrrrrr}
\toprule
Seed & Readout & Clean & Cache & $+0$ & $+1$ & $+2$ & $+4$ & P8 & P16 \\
\midrule
9 & $32$ & 99.80 & 99.80 & 3.91 & 12.30 & 96.09 & 99.80 & 7.42 & 4.30 \\
9 & $33$ & 100.00 & 99.80 & 3.91 & 9.18 & 93.75 & 100.00 & 9.38 & 4.30 \\
9 & $64$ & 100.00 & 99.80 & 3.91 & 9.18 & 90.23 & 100.00 & 7.81 & 5.08 \\
9 & Sched. & 99.80 & 99.80 & 3.91 & 9.18 & 90.23 & 100.00 & 7.42 & 4.30 \\
\midrule
10 & $32$ & 100.00 & 100.00 & 13.28 & 12.89 & 97.66 & 92.77 & 13.09 & 12.70 \\
10 & $33$ & 100.00 & 100.00 & 13.28 & 13.28 & 73.44 & 97.85 & 14.26 & 12.89 \\
10 & $64$ & 93.95 & 100.00 & 13.28 & 13.28 & 72.27 & 99.41 & 24.22 & 23.05 \\
10 & Sched. & 100.00 & 100.00 & 13.28 & 13.28 & 72.27 & 99.41 & 13.09 & 12.70 \\
\midrule
11 & $32$ & 100.00 & 100.00 & 1.56 & 53.12 & 100.00 & 100.00 & 1.76 & 1.76 \\
11 & $33$ & 100.00 & 100.00 & 1.56 & 38.09 & 100.00 & 100.00 & 1.76 & 1.56 \\
11 & $64$ & 99.22 & 100.00 & 1.56 & 38.09 & 99.22 & 100.00 & 5.66 & 1.56 \\
11 & Sched. & 100.00 & 100.00 & 1.56 & 38.09 & 99.22 & 100.00 & 1.76 & 1.76 \\
\bottomrule
\end{tabular}

\end{table}

\begin{table}[p]
 \centering
 \small
 \setlength{\tabcolsep}{4pt}
 \caption{\textbf{Every \sampledprefix{} whole-trajectory score under
 freezing.} Percent of 512 H32 examples, stream 20260924. Numerical
 readouts require all positions to be correct simultaneously.
 ``Sched.'' instead reads each position at its own $k+d$ for $+d$, or
 at $k$ for clean/cache/P8/P16, on that arm's trajectory.}
 \label{tab:control-freeze-sampled-trajectory}
\begin{tabular}{lrrrrrrrrr}
\toprule
Seed & Readout & Clean & Cache & $+0$ & $+1$ & $+2$ & $+4$ & P8 & P16 \\
\midrule
9 & $32$ & 100.00 & 100.00 & 0.00 & 100.00 & 100.00 & 100.00 & 0.00 & 0.00 \\
9 & $33$ & 100.00 & 100.00 & 0.00 & 100.00 & 100.00 & 100.00 & 0.00 & 0.00 \\
9 & $64$ & 100.00 & 100.00 & 0.00 & 100.00 & 100.00 & 100.00 & 0.00 & 0.00 \\
9 & Sched. & 100.00 & 100.00 & 0.00 & 100.00 & 100.00 & 100.00 & 0.00 & 0.00 \\
\midrule
10 & $32$ & 100.00 & 100.00 & 0.00 & 1.56 & 100.00 & 100.00 & 0.00 & 0.00 \\
10 & $33$ & 100.00 & 100.00 & 0.00 & 1.56 & 100.00 & 100.00 & 0.00 & 0.00 \\
10 & $64$ & 95.70 & 100.00 & 0.00 & 1.56 & 100.00 & 100.00 & 0.00 & 0.00 \\
10 & Sched. & 100.00 & 100.00 & 0.00 & 1.56 & 100.00 & 100.00 & 0.00 & 0.00 \\
\midrule
11 & $32$ & 100.00 & 100.00 & 0.00 & 0.39 & 99.22 & 100.00 & 0.00 & 0.00 \\
11 & $33$ & 100.00 & 100.00 & 0.00 & 0.39 & 98.83 & 100.00 & 0.00 & 0.00 \\
11 & $64$ & 94.14 & 100.00 & 0.00 & 0.39 & 98.63 & 100.00 & 0.00 & 0.00 \\
11 & Sched. & 100.00 & 100.00 & 0.00 & 0.39 & 98.63 & 100.00 & 0.00 & 0.00 \\
\bottomrule
\end{tabular}

\end{table}

\begin{table}[p]
 \centering
 \small
 \setlength{\tabcolsep}{4pt}
 \caption{\textbf{Every \matchedexit{} whole-trajectory score under
 freezing.} Same readout definitions as
 Table~\ref{tab:control-freeze-sampled-trajectory}; 512 H32 examples per
 model, stream 20260924. Clean scheduled exactness and common-loop
 exactness differ because clean position states continue to evolve.}
 \label{tab:control-freeze-matched-trajectory}
\begin{tabular}{lrrrrrrrrr}
\toprule
Seed & Readout & Clean & Cache & $+0$ & $+1$ & $+2$ & $+4$ & P8 & P16 \\
\midrule
9 & $32$ & 99.02 & 94.92 & 0.00 & 0.00 & 85.74 & 99.80 & 0.00 & 0.00 \\
9 & $33$ & 99.22 & 94.92 & 0.00 & 0.00 & 83.40 & 100.00 & 0.00 & 0.00 \\
9 & $64$ & 71.88 & 94.92 & 0.00 & 0.00 & 82.62 & 100.00 & 0.00 & 0.00 \\
9 & Sched. & 94.92 & 94.92 & 0.00 & 0.00 & 82.62 & 100.00 & 0.00 & 0.00 \\
\midrule
10 & $32$ & 32.03 & 99.80 & 0.00 & 0.00 & 14.06 & 91.99 & 0.00 & 0.00 \\
10 & $33$ & 27.54 & 99.80 & 0.00 & 0.00 & 13.28 & 97.66 & 0.00 & 0.00 \\
10 & $64$ & 0.20 & 99.80 & 0.00 & 0.00 & 12.30 & 99.22 & 0.00 & 0.00 \\
10 & Sched. & 99.80 & 99.80 & 0.00 & 0.00 & 12.30 & 99.22 & 0.00 & 0.00 \\
\midrule
11 & $32$ & 73.83 & 100.00 & 0.00 & 0.00 & 25.00 & 100.00 & 0.00 & 0.00 \\
11 & $33$ & 72.07 & 100.00 & 0.00 & 0.00 & 24.61 & 100.00 & 0.00 & 0.00 \\
11 & $64$ & 8.79 & 100.00 & 0.00 & 0.00 & 24.61 & 100.00 & 0.00 & 0.00 \\
11 & Sched. & 100.00 & 100.00 & 0.00 & 0.00 & 24.61 & 100.00 & 0.00 & 0.00 \\
\bottomrule
\end{tabular}

\end{table}

\begin{table}[t]
 \centering
 \small
 \caption{\textbf{Exactness of the never-frozen suffix.} Percent of all
 512 H32 examples per model, stream 20260924. Freezing prefix 1--8
 leaves suffix 9--32 live; freezing 1--16 leaves 17--32 live. Every
 position in that suffix must have both labels correct. The endpoint
 scores for these same arms appear in
 Tables~\ref{tab:control-freeze-sampled-endpoint}
 and~\ref{tab:control-freeze-matched-endpoint}.}
 \label{tab:control-freeze-prefix-suffix}
\begin{tabular}{lrrrrr}
\toprule
Condition & Seed & Frozen prefix & $r=32$ & $r=33$ & $r=64$ \\
\midrule
\sampledprefix & 9 & 8 & 0.00 & 0.00 & 0.00 \\
\sampledprefix & 9 & 16 & 0.00 & 0.00 & 0.00 \\
\sampledprefix & 10 & 8 & 0.00 & 0.00 & 0.00 \\
\sampledprefix & 10 & 16 & 0.00 & 0.00 & 0.00 \\
\sampledprefix & 11 & 8 & 0.00 & 0.00 & 0.00 \\
\sampledprefix & 11 & 16 & 0.00 & 0.00 & 0.00 \\
\matchedexit & 9 & 8 & 0.00 & 0.20 & 0.00 \\
\matchedexit & 9 & 16 & 0.00 & 0.00 & 0.00 \\
\matchedexit & 10 & 8 & 0.00 & 0.00 & 0.00 \\
\matchedexit & 10 & 16 & 0.00 & 0.00 & 0.00 \\
\matchedexit & 11 & 8 & 0.00 & 0.00 & 0.00 \\
\matchedexit & 11 & 16 & 0.00 & 0.00 & 0.00 \\
\bottomrule
\end{tabular}

\end{table}

\subsection{External maze network}
We evaluated the released \texttt{dt\_net\_recall\_2d} checkpoint (width 128,
trained on $9\times9$ mazes for 30 iterations with the progressive loss of
\citealp{bansal2022endtoend}) without modification, on 32 pre-selected mazes
per size for iterations 1--120. The recurrence was replayed by an external
loop that reproduces the released forward pass exactly; a never-triggering
hook reproduced the clean arm, and every arm matched clean before its first
capture. Because there is no per-position frontier, a pixel's capture time
is its \emph{answer arrival}: the last iteration at which its clean
prediction changes (mean 6.2 at $9\times9$, 19.7 at $33\times33$).
Table~\ref{tab:maze-freeze} gives whole-maze solved counts.

\begin{table}[h]
 \centering
 \small
 \caption{Solved mazes out of 32 for the released recall network. ``Capture
 $+d$'' freezes each pixel $d$ iterations after its answer arrival. Freezing
 every pixel at iteration 30 is harmless where the maze is already solved and
 fatal where computation is still running.}
 \label{tab:maze-freeze}
 \begin{tabular}{lrrr}
 \toprule
 Arm & $9\times9$ @30 & $9\times9$ @120 & $33\times33$ @120 \\
 \midrule
 Clean & 32 & 32 & 19 \\
 Answer cache (state untouched) & 32 & 32 & 19 \\
 Capture $+0$ & 0 & 0 & 0 \\
 Capture $+1$ & 3 & 3 & 0 \\
 Capture $+2$ & 28 & 28 & 2 \\
 Capture $+4$ & 32 & 32 & 18 \\
 All pixels at iteration 30 & 32 & 32 & 0 \\
 \bottomrule
 \end{tabular}
\end{table}

The clean network solves every $9\times9$ maze from iteration 25 through
120, so it does not forget answers, yet freezing at answer arrival solves
none. The limits are substantial. There is one checkpoint, no training
seeds, and 32 mazes per size. Architecture, task, and objective differ from
our models at once, so the check does not isolate any of them. The
network's output head is a $3\times3$ convolution stack, so holding a
pixel's state does not pin its prediction (2.9\% of frozen pixels drift at
capture $+0$, 0.3\% at $+4$, $9\times9$); its frozen-readout quantities are
not comparable with those of our per-position readout. The
$33\times33$ size lies outside the training distribution and its clean
accuracy is still rising at iteration 120, so those comparisons are
descriptive.

\clearpage
\section{Local recovery of the first natural failure}
\label{app:natural}

\paragraph{Protocol.}
The evaluation uses the step-132000 \finalonly\ checkpoints (seeds 9--14,
$d=128$; stream 20260909) and three \finalonly\ models with $d=512$ and
four layers per block (seeds 9--11; stream 20260923), with $H=32$ and 512
inputs per model. The primary cohort $E$ is fixed before any intervention:
inputs whose endpoint is jointly correct at loop 32 and jointly wrong at
loop 33. After loop 32, the listed positions are replaced by the model's own
vectors from an earlier loop of the same input, and loop 33 runs without
further intervention. Every batch passed a whole-state replay check, a
pre-patch invariance check, and a same-phase sham. The six \allprefix\
models have no member of $E$, because their loop-33 endpoint accuracy is
100\%; they have an empty rescue denominator, not a zero rescue rate.

\begin{table}[!htbp]
 \centering
 \small
 \setlength{\tabcolsep}{4pt}
 \caption{Recovered members of $E$ at loop 33. Columns: position 31 reset to
 its loop-31 vector (31), position 32 reset to its loop-31 vector
 (32), both, position 30 reset to its loop-31 vector (adjacent),
 position 32 reset to its loop-30 vector (earlier), and a same-phase
 self-replacement of both positions (sham). The last column counts inputs
 correct at both loops that each arm breaks at loop 33, for 31 /
 32 / both.}
 \label{tab:natural}
 \begin{tabular}{rrrrrrrrrc}
 \toprule
 $d$ & Seed & $|E|$ & 31 & 32 & Both & Adjacent & Earlier & Sham & Broken \\
 \midrule
 $128$ & 9 & 224 & 165 & 61 & 224 & 0 & 52 & 0 & 1 / 148 / 0 \\
 $128$ & 10 & 136 & 95 & 4 & 136 & 18 & 12 & 0 & 2 / 201 / 0 \\
 $128$ & 11 & 248 & 195 & 53 & 248 & 15 & 46 & 0 & 10 / 130 / 0 \\
 $128$ & 12 & 30 & 18 & 3 & 30 & 5 & 2 & 0 & 90 / 350 / 0 \\
 $128$ & 13 & 59 & 54 & 13 & 59 & 13 & 11 & 0 & 32 / 291 / 0 \\
 $128$ & 14 & 74 & 58 & 9 & 74 & 2 & 10 & 0 & 2 / 242 / 0 \\
 \midrule
 $512$ & 9 & 22 & 21 & 2 & 22 & 0 & 4 & 0 & 0 / 301 / 0 \\
 $512$ & 10 & 20 & 16 & 20 & 20 & 4 & 9 & 0 & 0 / 0 / 0 \\
 $512$ & 11 & 37 & 4 & 37 & 37 & 2 & 0 & 0 & 34 / 0 / 0 \\
 \bottomrule
 \end{tabular}
\end{table}

\paragraph{Results and limits.}
Pooled over the $d=128$ models, the position-31, position-32, and combined resets
recover 585, 143, and 771 of 771 members of $E$. Loop-33 accuracy on all
inputs is 89.5\%, 35.2\%, and 100\%, against 74.9\% clean. For $d=512$ the
counts are 41, 59, and 79 of 79, with all-input accuracy of 95.3\%, 79.1\%,
and 100\% against 94.9\% clean. Single resets are not ranked: the position-31
reset leads in $d=512$ seed 9, the position-32 reset in seeds 10 and 11, and a position-32
reset alone breaks between 130 and 350 correct inputs in every $d=128$
model. The factorial interaction varies in sign across seeds, so we do not
claim superadditivity.

The combined reset leaves positions 1--30 one loop ahead but otherwise
approximately replays loop 32 at the endpoint, which was correct. Its
success therefore localizes the first failure to the last two positions; it
does not show that a restored state persists. Without a second
intervention, $d=128$ all-input accuracy at loop 34 is 63.7\% after the
combined reset versus 65.5\% clean, whereas for $d=512$ it is 93.4\% versus
76.8\%. We report the relapse as a $d=128$ observation only. The experiment
identifies joint sufficiency at one horizon and loop; it does not identify a
circuit, show that either reset is necessary, or establish a repair method.

\end{document}